\documentclass[sigconf, nonacm]{acmart}

\usepackage{hyperref}       
\usepackage{url}            
\usepackage{booktabs}       
\usepackage{amsfonts}       
\usepackage{nicefrac}       
\usepackage{microtype}      
\usepackage{cleveref}

\usepackage{graphicx}
\usepackage{doi}
\usepackage{upquote}
\usepackage[colorinlistoftodos]{todonotes}
\usepackage{siunitx}
\usepackage{subcaption}
\usepackage[table]{xcolor}  
\usepackage{multirow}
\usepackage{enumitem}
\usepackage{tikz}
\usetikzlibrary{arrows.meta}
\usepackage{array}
\newcolumntype{C}[1]{>{\centering\arraybackslash}p{#1}}
\usepackage{listings}
\lstdefinestyle{prompt}{
    gobble=4,
    showstringspaces=false,
    basicstyle=\ttfamily\small\color{blue!50!black},
    breakindent=0pt,
    breakatwhitespace=true,
    breakautoindent=false,
}
\lstdefinelanguage{Cypher}{
  morekeywords=[1]{ 
    MATCH, OPTIONAL, RETURN, WHERE, WITH, UNWIND, CREATE, MERGE,
    DELETE, DETACH, SET, REMOVE, ORDER, BY, SKIP, LIMIT, UNION,
    ALL, CALL, YIELD, USE, LOAD, CSV, FROM, FOREACH, CASE, WHEN,
    THEN, ELSE, END, AS, ON, CONSTRAINT, INDEX, DROP, SHOW
  },
  morekeywords=[2]{ 
    AND, OR, NOT, XOR, IN, IS, NULL, STARTS, ENDS, CONTAINS,
    DISTINCT, ASC, DESC, EXISTS, ANY, NONE, SINGLE
  },
  morekeywords=[3]{ 
    count, collect, sum, avg, min, max, size, length, type, labels,
    keys, properties, nodes, relationships, id, elementId, toString,
    toInteger, toFloat, toBoolean, coalesce, head, tail, last, range,
    reduce, abs, ceil, floor, round, rand, timestamp, datetime, date,
    time, duration, point, split, replace, substring, trim, toLower,
    toUpper, left, right
  },
  morekeywords=[4]{true, false, TRUE, FALSE}, 
  sensitive=true,
  morestring=[b]', 
  morestring=[b]", 
  morecomment=[l]{//}, 
  morecomment=[s]{/*}{*/}, 
  alsoletter={_},
}
\lstdefinestyle{cypher}{
  language=Cypher,
  basicstyle=\ttfamily,
  keywordstyle=[1]\color{blue!70!black},    
  keywordstyle=[2]\color{purple!70!black},  
  keywordstyle=[3]\color{teal!70!black},    
  keywordstyle=[4]\color{red!60!black},     
  stringstyle=\color{green!50!black},
  commentstyle=\color{gray},
  showstringspaces=false,
  breaklines=true,
  breakatwhitespace=true,
  breakautoindent=false,
  breakindent=0pt,
}

\newcommand{\rotv}[1]{\rotatebox{90}{#1}}

\begin{document}
\title{Reducing Hallucinations in Complex Question Answering using Simple Graph-based
  Retrieval-Augmented Generation (long~version)}





\author{Christopher J.~Wedge, Joshua Stutter, Danny Dixon, Jacek Ca\l{}a}
\orcid{0000-0002-8322-4370}
\affiliation{%
  \institution{National Innovation Centre for Data, Newcastle University}
  \streetaddress{3 Science Square}
  \city{Newcastle upon Tyne}
  \country{UK}
}
\email{{Chris.Wedge, Joshua.Stutter, D.J.Dixon2, Jacek.Cala}@ncl.ac.uk}

\begin{abstract}
Large language models (LLMs) have fundamentally transformed the landscape of Natural Language Processing (NLP), although they remain susceptible to errors.
Retrieval-augmented generation (RAG) systems have emerged as a common deployment scenario seeking to both avoid the well known risk of the LLM ``hallucinating'' information, and to enable reasoning and question answering over proprietary information that the LLM did not have access to during training without resorting to expensive model fine-tuning.

In this work, we explore the idea of using a lightweight graph structure with a relatively simple graph schema, to support the RAG subsystem via a dedicated toolset.
We design an agentic system with a variety of vector search and graph query tools operating over a structured dataset based on a curated subset of English Wikipedia articles, and evaluate its performance on questions from MoNaCo, a challenging Wikipedia based benchmark of complex question answering (QA) tasks.

Our results show that the introduction of graph-based tools can significantly increase the precision and recall of factual correctness, can halve the number of hallucinated answers, and achieves the highest fine-grained truthfulness score among the three evaluated scenarios.
All this with a modest increase in token usage.
\end{abstract}

\maketitle

\begingroup
\renewcommand\thefootnote{}\footnote{\noindent
This work is licensed under the Creative Commons BY-NC-SA 4.0 International License. Visit \url{https://creativecommons.org/licenses/by-nc-sa/4.0/} to view a copy of this license. For any use beyond those covered by this license, obtain permission by emailing the authors. Copyright is held by the owner/author(s). \\
\raggedright 
}\addtocounter{footnote}{-1}\endgroup


\section{Introduction}

Since the advent of the transformer architecture many competing Large language models (LLMs) have been developed
with strong capabilities in understanding, reasoning, summarisation, and question answering.
Their ability to generate coherent and contextually relevant responses has led to widespread adoption in many domains.
Despite these advances, LLMs and LLM-based systems remain prone to a variety of failure modes.

Type I (false positive) and Type II (false negative) errors are not only core
concepts in statistical inference, but also provide a useful framework for
describing fundamental failure modes in statistical, analytical, and machine-learning models, including LLMs.
In the case of LLMs and LLM-based systems, the ramifications of these failures can be severe and 
multifaceted.  False positives may manifest as hallucinated content, retrieval of the incorrect
context, or the use of inappropriate tools, whereas false negatives may lead to omissions,
missed relevant context, or failure to invoke the correct tools.
In practice, such errors often occur simultaneously in multiple
stages of system operation~\citep{dziri23faith}.
Therefore, any method that improves the quality of a model or a model-based system is critical, insofar as
it reduces both types of error simultaneously rather than merely shifting
performance along the precision-recall tradeoff. This consideration becomes even
more vital for complex question answering problems, such as multi-hop (MHQA),
cross-document (CDQA) and multi-entity question answering
(MEQA)~\citep{yang18hotpot,caciularu21cdlm,lin25mebench}, where
knowledge from multiple sources must be retrieved, aggregated, and reasoned over
in order to provide an answer.

As an example, consider the following question:
\begin{quote}
\emph{Can you name all the battles between the Dutch and English in the First, Second and Third Anglo-Dutch Wars, and list the victor of each battle?}
\end{quote}
To answer this question, one needs to perform a sophisticated retrieval and
reasoning process.  In fact, it requires multi-entity and multi-hop reasoning,
and cross-document access, all at once~\citep{trivedi2022musique}.

First, as with MEQA, the process must operate on a set of entities (in this case
nations, wars, battles and victors) and needs to perform per-entity analysis and
filtering.  Second, as with MHQA, it requires hierarchical traversal, repeated
relational expansion and nested decomposition.  Finally, as with CDQA, it needs
to fetch information from multiple sources: war documents, battle descriptions,
and historical records.
A single document is unlikely to answer this question; the information required
is typically dispersed across dozens of documents.

These types of questions pose a significant challenge to current state-of-the-art LLM-based
systems.  There are multiple reasons for this, but in this work we focus on one aspect: Retrieval-Augmented Generation (RAG)~\citep{lewis_2020_retrieval}.
RAG is a common element of many LLM-based systems, specifically those used for
document processing, analytics, and question answering.
In this context, we address complex QA problems, like the one presented above, using a unified vector and graph database with a series of pre-defined tools to improve retrieval from a knowledge base (KB).
We compare our solution, vector+graph RAG, with simple vector RAG and zero-shot approaches.
Using LLM agents prompted for safe refusal, that is models explicitly instructed to state `unknown' when the necessary information could not be found in the provided KB, our system more than halved the proportion of complex questions that the agent refused to answer.
At the same time, it improved the ratio of correct over incorrect answers compared to the zero-shot approach.
Thus, it shows a promising direction of research in which graphs are coupled with more traditional vector-based retrieval methods.

The use of knowledge retrieval systems (RAG) to extend the capability of LLM agents is the preferred method by which general-purpose LLMs are applied to reason over specialised and\,/\,or proprietary knowledge~\citep{fan24survey}.
By augmenting a user query with data queried by a 
knowledge retrieval system -- data which may not be in the LLM's training data
-- the LLM can answer the question using in-context learning, extending its 
capabilities and reducing hallucinations~\citep{niu24ragtruth}.

Whilst early RAG systems consisted of a single retrieval-then-answer 
pipeline~\citep{gao24retrieval}, it is now typically expected that RAG systems
should be agentic for all but the simplest queries, so as to allow the LLM to
practise question decomposition, reflection, and follow-up or validation queries
using Chain of Thought (CoT) reasoning. 
Limiting ourselves to systems querying a defined KB (rather than Internet search), the most
commonly-used tool within agentic RAG is vector search, where a vector database
of text chunks from source documents is queried by an embedding  model to assess
semantic similarity~\citep{lee19latent,brown25systematic}.
Beyond this, a variety of other tools have been developed that purport to
improve retrieval in terms of its accuracy and\,/\,or efficiency.

Among these new retrieval methodologies, those that employ graphs are gaining popularity.
Rather than operating over unstructured,
chunked text documents, a knowledge graph (KG) is either harnessed directly or
created using an LLM from a set of related documents (commonly known as 
\textit{GraphRAG}~\citep{edge2025localglobalgraphrag}).
However, a major challenge with graph-based solutions is the creation of a
useful KG. On the one hand, KGs with rich structures are potentially desirable, 
but they necessitate a complex schema and thus risk filling the fixed context
window of an LLM~\citep{chakraborty24multihop}. On the other hand, small schemas are more easily digestible by an LLM, but may not faithfully and comprehensively
capture the true status of the KB.

In this work, we explore the latter idea, i.e.\ using a lightweight graph structure, with a relatively simple graph schema, to support the RAG subsystem via a dedicated
toolset.
We adopt this very practical approach in which a KB is constructed from semi-structured documents, including only a high-level structure of document titles, section titles, and constituent paragraphs, with links to other paragraphs or documents.
This scenario is well suited to many real-world KBs that comprise various document types but lack a sophisticated and\,/\,or rich KG representation, such as those proposed by other work~\citep{dong14knowledge}.

The tools we developed use the \textit{Neo4j} graph database engine
and its accompanying \textit{Cypher} query
language\footnote{\url{https://neo4j.com}} to traverse the graph of Wikipedia
articles, article sections and section paragraphs.  We compare our graph-based
approach with a typical vector-based RAG system on a set of sophisticated queries that
require gathering and reasoning over information from multiple sources
(articles, sections, and paragraphs).

Although work has been done to evaluate vector RAG against \textit{GraphRAG}
methods~\citep{
  kosten2023spider, liu2024spinach, feng25rgrkbqa, xiao2025graphragbench%
}, in this work we are concerned instead with evaluating vector RAG against knowledge base question answering (KBQA) from the perspective of unstructured and semi-structured data. There is as yet no research that directly evaluates vector RAG against KBQA on structured data. To address this deficiency, we aim to answer the following research question:
\begin{quote}
\textit{Do queries over structured knowledge bases, such as knowledge graphs, improve the performance of complex QA over unstructured RAG?}
\end{quote}
We design an agentic system with a variety of vector search and graph query
tools operating over a structured dataset based on a curated subset of English 
Wikipedia articles, and evaluate its performance on a challenging Wikipedia QA 
benchmark (MoNaCo, \emph{vide infra}).
Our experiments measure both answer accuracy and token usage, with the objective of a strong system being to maximise answer accuracy while minimising token usage.

Additionally, we compare both solutions against a zero-shot approach that relies purely on the knowledge acquired during model training.
Our results show that the introduction of graph-based RAG significantly reduces hallucinated content and improves truthfulness scores, all with only a modest increase in token usage compared to the more conventional vector RAG approach, and using a lightweight and easy to construct graph knowledge base.

In summary, our main contributions are as follows. We:
\begin{itemize}
    \item Identify a complex-QA benchmark suitable for comparing structured and
    unstructured retrieval.
    \item Design a unified \textit{Cypher}-based toolset that affords efficient navigation over a knowledge graph.
    \item Conduct extensive experiments on the efficacy of RAG and KBQA tools.
    \item Release an evaluation KG based on a significant proportion of English Wikipedia, suitable for both RAG and KBQA.\footnote{\url{https://github.com/NICD-UK/graph-based-rag-qa/}; to be shared upon publication.}
\end{itemize}

\section{Background and Scope}

As has recently been observed, the expectations of LLM-based and agentic
systems are increasing beyond simple question answering problems~\citep{fan24survey}.
Indeed,
real-world queries often require retrieving information from multiple sources,
summarising and reasoning before an answer can be produced.  In this context,
our reliance on systems that may hallucinate their answers, even if only at some
of the substages of the whole answering pipeline, is problematic and undermines
trustworthiness of these systems.  Therefore, the need to design and develop
solutions that can reduce hallucination (confabulation) and improve trust are
critical.

\begin{figure*}
    \centering
    \includegraphics[width=.62\linewidth]{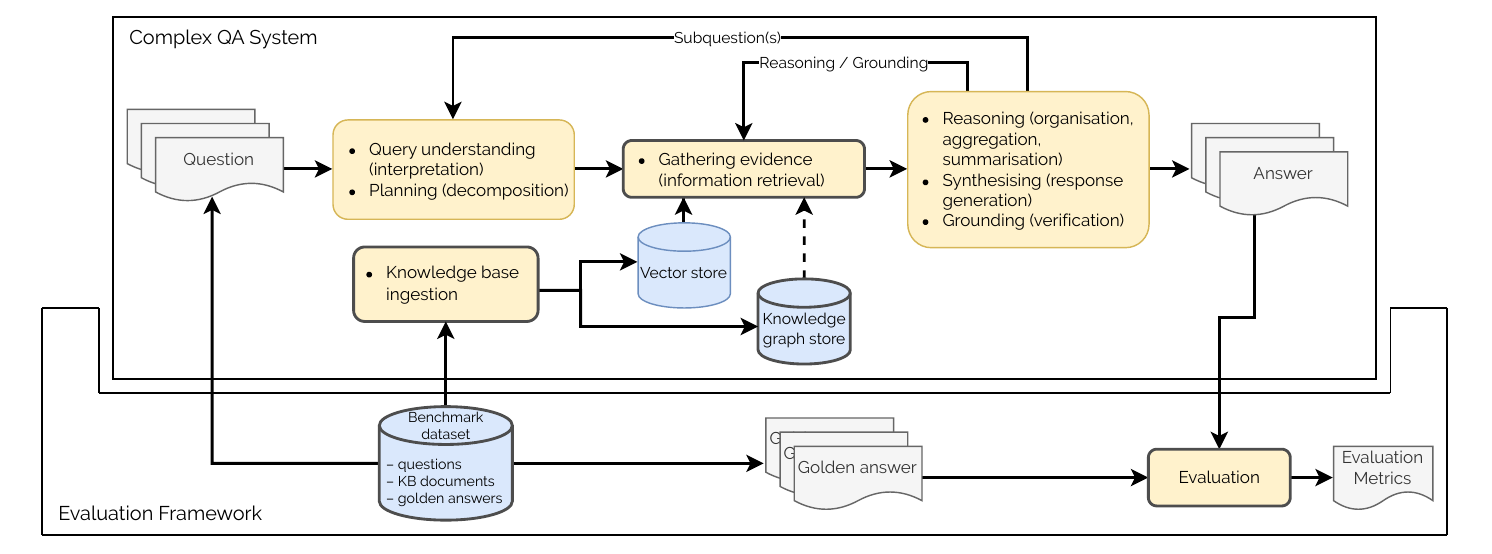}
    \caption{
      A Graph-based Question Answering pipeline embedded within an evaluation
      framework. Highlighted in black are items of our focus.
    }
    \label{fig:qa-pipeline}
\end{figure*}

Recently, there have been various attempts to address this problem.  They
revolve around improving four main aspects: query understanding and
decomposition, evidence gathering (i.e.\ information retrieval), reasoning, and
response synthesis and grounding (cf.~\Cref{fig:qa-pipeline}).

In this work we focus on the second step -- information retrieval -- and aim to
verify the extent to which the introduction of a knowledge graph store with a
simple document graph structure within a complex QA system can improve overall
system performance.  To accomplish this, we embed our QA system within an
evaluation framework that includes a benchmark dataset and an evaluation step.
Our goal is to evaluate the end-to-end factual correctness and truthfulness of the QA system.
Our comparison is deliberately scoped to generic vector RAG versus vector+graph
RAG over a simple document graph; more sophisticated \textit{GraphRAG}~\citep{edge2025localglobalgraphrag} approaches and
richer, LLM-constructed KG representations are beyond the scope of this study.
Our simple graph is
intended as an \emph{enabler} of structure-aware retrieval rather than a
competing heavyweight knowledge graph.

\subsection{Information Retrieval}

In the domain of information retrieval for LLM agents, there are two main approaches: parametric and non-parametric~\citep{asai23retrieval}.
Parametric methods, such as fine-tuning, are beneficial for encoding knowledge directly within a model's weights, but they can be incredibly resource intensive and are inflexible to new knowledge~\citep{soudani24finetuning}.
Non-parametric solutions, such as RAG, on the other hand, are used directly with base models.
By augmenting a user query with data returned by a knowledge retrieval system, i.e. the data that may not be in the LLM's training data, the LLM can answer the question using in-context learning, extending its capabilities and reducing hallucinations~\citep{niu24ragtruth}.

Several paradigms have emerged within state-of-the-art RAG systems, to effect this application of knowledge retrieval to LLMs.
Vector search is still the most
common method, as it can be applied directly to unstructured text,
such as a corpus of PDF documents on a topic.  As a result, most RAG benchmarks
and evaluations begin from the assumption that the initial data will be of an
unstructured form~\citep{han25retrieval}.  For natively structured data,
text-to-SQL~\citep{deng25reforce,li23can} or 
text-to-SPARQL~\citep{soru17sparql,kovriguina23sparqlgen} are common techniques,
where the LLM is tasked with directly querying a relational or graph database respectively.

For the case of structured data querying over graphs, methods either:
\begin{itemize}
  \item use graph metrics, such as community detection, alongside vector retrieval, to improve results by retrieving small 
  subgraphs~\citep{edge2025localglobalgraphrag},
  \item verbalise entities and relationships into text for
  embedding \citep{baek23direct}, or
  \item generate structured queries, e.g. in SPARQL or \textit{Cypher}, that are then
  executed directly against the graph~\citep{feng25cypherbench}.
\end{itemize}
Here, we propose an alternative approach.
We expose the agent to specific tools, backed by handwritten queries, which enable flexible structured data retrieval while, at the same time, relieve the agent from the task of generating valid data queries -- a potential source of failure and security vulnerability.

\subsection{Benchmark dataset}

An external knowledge base offers three advantages over relying on an LLM's
parametric memory: it can be updated continuously rather than being frozen at
training time~\citep{petroni19language,lewis_2020_retrieval}; it supplies
domain-specific or proprietary knowledge that the model
lacks~\citep{kandpal23large}; and it sidesteps the degraded performance of LLMs
on long or complex inputs, since even large context windows are usably shorter
than advertised~\citep{hsieh2024ruler}.  Regardless of context size or parameter
count, there will always be datasets too large or specialised for an LLM to
reason over unaided.

Typical vector RAG and automatically-constructed 
\textit{GraphRAG} both ingest
\emph{unstructured} documents.  By contrast, a graph database operates on
already-structured data, and its key advantage is a controlled schema: rather
than \emph{searching} for semantically similar text, it lets us \emph{query}
structure directly.  Using this lens, we reviewed recent datasets and benchmarks
to find one usable in a knowledge-graph RAG (KG RAG) system, employing the
following criteria:

\begin{enumerate}
  \item Large dataset (>10,000 records); for a realistic scenario in which the KB does not fit into the LLM context window.
  \item End-to-end QA; in contrast to retrieval only or generation only evaluation.
  \item Data appropriate for both structured and unstructured
  retrieval; for fair comparison between simple vector-based RAG and KG RAG.
  \item Multi-hop, multi-entity reasoning; to tackle the challenging task of complex QA which the current LLM-based systems struggle to solve.
\end{enumerate}
A search within the literature for benchmark datasets for RAG, especially
graph RAG, yielded many candidates.
A summary of this survey is presented in Appendix~A.

In the end we selected the \emph{MoNaCo} benchmark dataset~\citep{wolfson_m_2026}.
Although it does not directly include a structured KB we built a simple knowledge graph to provide a structured index to the source documents (Wikipedia articles) necessary as context.
This benchmark dataset, contains 1315 complex questions, one of which was shown earlier in the introduction.
They are human written in natural language rather than being LLM-generated as in some alternatives.
The answer to each question is a result to be determined by combining or reasoning over multiple pieces of information. The published, yet preliminary, studies of this dataset show poor performance by modern LLMs in a zero-shot or basic RAG context, which requires the use of LLM reasoning approaches~\citep{wolfson_m_2026}.
Within an agentic vector RAG implementation, this might require multiple queries to retrieve the necessary information, but we postulate that using a KG RAG system it may be possible to retrieve information directly via a suitable graph query.

All this provides a very realistic scenario.
It is relevant to answering questions from a store of documents based on a simple and easy to create graph rather than a more comprehensive graph database.
The latter might not be present in many application settings.
Although this dataset is less than ideal in that the QA pairs are based on publicly available Wikipedia data, the complexity of questions, and the reported failure of zero-shot LLMs to answer these accurately without a retrieval subsystem~\citep{wolfson_m_2026}, ensure that the agent cannot rely solely on parametric knowledge.

\subsection{Evaluation}\label{sec:eval}

The decision on the most appropriate RAG evaluation metrics must take into consideration the benchmark dataset in use, to ensure that the required information is available to compute the chosen metrics in both vector- and KG-RAG scenarios.
For definitions of key metrics see the two recent reviews of RAG evaluation methods in \citep{gan_2025_retrieval,yu_2025_evaluation}. 
Here, we consider which of the many available metrics are appropriate.

RAG-supported systems comprise separate retrieval and generation
components~\citep{lewis_2020_retrieval}.
As a result, any evaluation experiment has the option to focus on only one of the two individual components, or the
entire end-to-end pipeline~\citep{yu_2025_evaluation}.
For the results to be widely explainable, in this study we focus on metrics
that are relevant to the end-to-end RAG system.  We report a headline difference in overall performance when adding a KG to a RAG system, rather than focusing on
a specific sub-component.

\subsubsection{Traditional metrics.}\label{subsec:answer-metrics}
The simplest metrics for evaluating generated answers come from the domain of NLP and tend to work at a character level.
These include Exact Match, BLEU and ROUGE.
The challenge when applying such metrics to LLM output is that a correct answer, identical semantically but phrased differently to the reference answer, would be given a low score.
They are, therefore, ineffective when evaluating free-form LLM responses~\citep{gan_2025_retrieval, yang_2024_crag}.

\subsubsection{Embedding-based metrics.} Metrics based on semantic similarity are more appropriate for our purpose, as they seek to evaluate whether the meaning
of the generated and reference answers are the same.  A typical example would be
BERTScore, which generates embeddings from a BERT pre-trained transformer for
both generated and reference answers, allowing token level similarity to be
addressed~\citep{zhang20bertscore}.
Similar models include BART and RoBERTa~\citep{gan_2025_retrieval}.
These approaches have the advantage of capturing some semantic meaning, while still relying on a deterministic scoring method.
However, they are less flexible and give lower evaluation quality
than LLM prompting methods~\citep{gao_2025_llm}, and some exhibit specific blind-spots~\citep{he_2023_blind}.

\subsubsection{LLM-based evaluators.}\label{subsec:llm-based-eval}
The final major class of evaluation metrics to discuss are the so-called \emph{LLM-as-a-judge} methods.
These rely on prompting an LLM to evaluate some material from the RAG chain against predefined criteria.
This general approach can be used for a wide variety of different metrics assessing the RAG system at various stages; the required inputs to evaluate these are summarised in \Cref{tbl:evaluation_metrics}.

For example, the Ragas framework~\citep{ragas_2024} has \textit{context precision} and \textit{context recall} metrics, in which retrieved contexts are processed along with the user query and golden answer to
determine whether they are useful in answering that query.
The computation of these scores is achieved by decomposing the number of distinct claims within the contexts to assess precision and recall on a claim-by-claim basis, and the reference answer is used as a proxy for a golden context to allow evaluation even when a curated dataset containing the golden context is not available.

\begin{table}[tb]
\caption{
  Summary of different LLM-based RAG evaluation metrics with required
  inputs and score range; \underline{underlined} are metrics used in this work.
  \label{tbl:evaluation_metrics}
}
\centering
\footnotesize
\begin{tabular}{c|ccccc}
\toprule
            & User & Retrieved & LLM & Golden & Score \\
     Metric & Query & Context & Response & Answer & Range\\
\midrule
    \underline{Answer Relevancy} & $\bullet$ & & $\bullet$ & & 0 to 1\\
    Faithfulness & $\bullet$ & $\bullet$ & $\bullet$ & & 0 to 1\\
    Context Precision & $\bullet$ & $\bullet$ & & $\bullet$ & 0 to 1\\
    Context Recall  & $\bullet$ & $\bullet$ & & $\bullet$ & 0 to 1\\
    \underline{Factual Correctness} & & & $\bullet$ & $\bullet$ & 0 to 1\\
    \underline{CRAG} & $\bullet$ & & $\bullet$ & $\bullet$ & 1, 0 or -1 \\    
\bottomrule
\end{tabular}
\end{table}

Two LLM-based metrics commonly seen to be important for RAG systems are
\textit{faithfulness} and \textit{answer relevancy}.  Both of these metrics can
be computed using the user query and LLM-generated response without requiring a
golden answer, and were introduced by Ragas to address the specific challenge of
RAG evaluation when a golden answer is not available~\cite{es_ragas_2024}.
\textit{Answer relevancy} aims to measure alignment between the generated
response and the user query, by generating three additional queries from the
response and computing the average cosine similarity between the embedding of
these queries and the actual query.  \textit{Faithfulness} additionally uses the
retrieved contexts, and breaks down the individual claims in the LLM response to
check these for support from the context.  The ratio of supported claims to
total claims is then returned as a score from 0 to 1 (e.g.\ a faithfulness score
of 1 means that all claims in the answer are justified by the retrieved
context).

A potential challenge to the evaluation of faithfulness is that for
complex questions with extensive reasoning or aggregation over multiple context
chunks even correct answers will not have a direct corresponding claim in a
single context chunk, which may lead to errors in the evaluation.
As the MoNaCo benchmark does not provide specific golden context chunks directly, and prompting for metrics using context chunks is not developed for the scenario of reasoning and aggregating from multiple chunks, the context precision/recall and faithfulness metrics were deemed inappropriate and are not used in this study. 

A more general LLM-as-a-judge metric is the \textit{factual correctness} score
in Ragas, which is computed by a direct comparison between the generated and
reference answers.
In this case, the LLM is asked to decompose each input into a number of individual claims.  By assessing using an LLM judge whether
each claim in the response list matches a claim in the reference list and vice versa, claims are labelled as true positives, false positives or
false negatives, allowing the system to calculate precision, recall or F\textsubscript{1} score according to their usual definitions~\citep{ragas_2024}.
While score calculation is done analytically, the claim decomposition and verification steps rely on LLM judgement.
This allows flexibility in answer interpretation to determine correctness without direct word-for-word text matching, but it does make the results non-deterministic.

\subsubsection{Custom metrics.}\label{subsubsec:custom_metrics}
While all of the metrics discussed above are commonly used, there are several
other useful metrics for RAG evaluation that are used for specific purposes.
One is the \textit{Comprehensive RAG Benchmark} (CRAG) score, introduced by~\citep{yang_2024_crag}.  A key observation made by the authors is that incorrect
or hallucinated answers are far worse than missing or refused answers, as they
can harm user confidence in the LLM system.  The CRAG paper therefore assigned
$+1$ to a correct answer, $0$ to a missing answer, and $-1$ to a hallucinated
answer.  For human evaluation, the scoring was extended to include not only $+1$
for a perfect answer but also $+0.5$ for an acceptable answer that contains
useful information but with some minor errors or omissions.  However, for auto
evaluation with an LLM judge, both $+1$ and $+0.5$ were combined into the
accurate class ($+1$).  With this scoring system, a value can be reported
for each answer in terms of whether it is accurate, missing, or hallucinated.
From these can be calculated the proportions of each answer class, as well as an
additional \textit{truthfulness} metric,
being the sum of all the scores given across a question batch (i.e. the number of accurate answers less the number of hallucinated answers).

\section{A Solution to Graph-based Complex Question Answering}

\begin{figure}
    \centering
    \includegraphics[width=.9\linewidth]{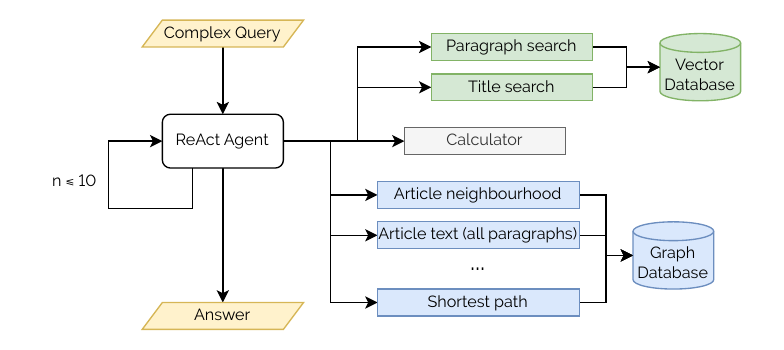}
    \caption{
      The architecture of our QA system with three types of tools: vector-based
      (green), graph-based (blue) and a simple calculator tool; middleware
      applies a soft limit of 10 iterations to the tool use loop.
    }
    \label{fig:architecture}
\end{figure}

To implement the QA pipeline presented earlier in \Cref{fig:qa-pipeline}, we developed a system with a single reasoning agent and a set of tools used to retrieve information from vector and graph databases.
\Cref{fig:architecture} shows the architecture of the proposed solution.

The reasoning agent was created using LangChain's \emph{create\_\-ag\-ent}
function,\footnote{\url{https://reference.langchain.com/python/langchain/agents/factory/create_agent}.} which provides tool calling capabilities.
We equipped the agent with three types of tools: vector-based, graph-based and a
simple calculation tool.

The agent was supplemented with standard middleware for tool call error
handling, retries with exponential back-off for LLM failures, and a run
limiter to prevent a never ending cycle of tool calls and limit token 
consumption.  Finally, the agent was given a JSON structured output schema such
that an answer, explanation for the answer, and references were produced
separately.

The database, both vector store and graph DB, were
populated with a semi-structured representation of a significant portion of
English Wikipedia articles included in the MoNaCo benchmark.

\subsection{MoNaCo reproducibility}

The question-answer pairs of MoNaCo were created by 24 Amazon
Mechanical Turk (AMT) crowdworkers over an unspecified period of time on live English Wikipedia~\citep{wolfson_m_2026}.
Although this is methodologically straightforward, it creates issues for reproducibility.
A key concern is that Wikipedia is constantly in flux: articles that AMT workers may have used to generate question-answer pairs may have since been altered, deleted, or more up-to-date articles created.
Therefore, in this study we used the benchmark against a snapshot version of the English Wikipedia from August 2025.
This was done after we conducted a data alignment experiment to ascertain which of the various snapshots of Wikipedia prior to the benchmark publication was capable of providing the most valid sources to perform the evaluation.

Each of the fetched snapshots, from January 2023 to August 2025, were scored according to the proportion of answerable questions, out of the MoNaCo total of 1315.
The August 2025 snapshot contained the sources necessary to answer 1207 ($91.8\%$) of the questions.
The 108 questions deemed unanswerable by this snapshot were dropped from our evaluation.

\subsection{Knowledge graph creation}\label{sec:graph_creation}

Each article within the snapshot consists of \emph{Wikitext}, a lightweight markup language used for authoring Wikipedia articles.
Using Wikitext markers, we parsed sections and paragraphs from each article as well as article redirects and articles linked to from each paragraph.
From these links, we also materialised links between articles themselves (mentions), which later aid path-finding within the dataset.

\begin{figure*}
	\centering
    \resizebox{.6\linewidth}{!}{%
        \begin{tikzpicture}[
            >={Stealth[length=2.5mm]},
            entity/.style={font=\Large,
                circle, draw=black, very thick, minimum size=1.9cm, inner sep=0pt
            },
            rel/.style={->, thick, draw=black!80},
            rellbl/.style={font=\Large, fill=white, inner sep=2pt,
            rounded corners=1pt, align=center},
            proplbl/.style={align=center, font=\Large},
            node distance=4cm,
            transform shape
        ]
            \node[
                entity,
                label={[proplbl]below:title (string)\\embedding (vector)}
            ] (article)   {Article};
            \node[
                entity,
                right=of article
            ] (section)   {Section};
            \node[
                entity,
                right=of section,
                label={[proplbl]below:text (string)}
            ] (paragraph) {Paragraph};
            \node[
                entity,
                right=of paragraph,
                label={[proplbl]below:text (string)\\embedding (vector)}
            ] (chunk)     {Chunk};
    
            \draw[rel] (article)   -- node[rellbl] {HAS\_SECTION}   (section);
            \draw[rel] (section)   -- node[rellbl] {HAS\_PARAGRAPH} (paragraph);
            \draw[rel] (paragraph) -- node[rellbl] {HAS\_CHUNK}     (chunk);
    
            \path (section)   edge[rel, loop, out=50, in=130, looseness=5]
                node[rellbl] {NEXT\_SECTION\,/\\PREVIOUS\_SECTION} (section);
            \path (paragraph) edge[rel, loop, out=50, in=130, looseness=5]
                node[rellbl] {NEXT\_PARAGRAPH\,/\\PREVIOUS\_PARAGRAPH} (paragraph);
            \path (chunk)     edge[rel, loop, out=50, in=130, looseness=5]
                node[rellbl] {NEXT\_CHUNK\,/\\PREVIOUS\_CHUNK} (chunk);
    
            \path (article) edge[rel, loop, out=50, in=130, looseness=5]
                node[rellbl] {MENTIONS} (article);
            \path (article) edge[rel, loop, out=140, in=220, looseness=6]
                node[rellbl] {REDIRECTS\_TO} (article);

            \draw[rel] (paragraph) to[bend left=24]
                node[rellbl, pos=0.5] {LINKS\_TO} (article);
        \end{tikzpicture}
    }
    \caption{Graph schema for the English Wikipedia dataset}
    \label{fig:schema}
\end{figure*}
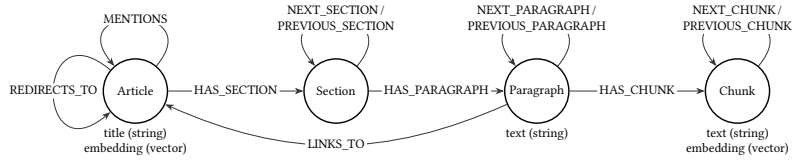

As we also wanted to evaluate vector RAG, paragraphs were further split into 500-character chunks, with a 40-character overlap, following the parameters recommended by~\citet{bratanic25essential}.
Preceeding and succeeding sections, paragraphs, and chunks were made into doubly-linked lists via the \texttt{NEXT\_*} /\texttt{PREVIOUS\_*} relationships.
Article nodes were given a \texttt{title} property, and the text of each paragraph was stored in a \texttt{text} property, which was then replicated onto the relevant chunk.
The resulting schema is shown in Figure~\ref{fig:schema}.

Using the filtered list of 1207 MoNaCo questions, a canonical list of 28,240 source articles was created.
The August 2025 snapshot was filtered to just these sources, to balance dataset size against information density.
Nodes outside of this filtered subset that were linked to from inside the subset were reduced to stubs: for example, an article not listed as a source by MoNaCo but linked to by a paragraph within a MoNaCo source was reduced to an empty article node of which there were 1,022,549.

The resulting KG comprised 5,771,867 nodes and 22,088,251 relationships.
Nearly half of all nodes were paragraph chunks, which were vectorised to enable vector search.
As such, an embedding model was selected based on its performance on the retrieval task of the MTEB
leaderboard~\citep{mteb26leaderboard}.

At the time of writing, the top-performing models on the MTEB leaderboard for retrieval were almost entirely
large embedding models with 8B parameters or more, such as
\textit{Qwen3-embedding}.
Despite demonstrating high performance, such
large models are unfeasible for this study, where the dataset within the chunk's
\texttt{text} attribute runs to over 300 million tokens, and so the embedding process would
take an unreasonably long time.
However, an outlier among the top retrieval
embedding models on the leaderboard was Microsoft's \textit{Harrier}
0.6B~\citep{microsoft26harrier}, which achieved top-5 performance on retrieval
tasks (70.75), similar to \textit{Qwen3-embedding} 8B (70.88), and top-10
performance overall (69.01), despite being a much smaller and faster model.

Apart from chunk embeddings, an embedding was also created for each article title, and the resulting dataset was imported to the Neo4j database.
Vector indices were created for each chunk's text embedding and article title embedding, so that the graph could be queried directly via \textit{Cypher}, or article titles / chunk texts surfaced via vector search.
The resulting dataset could be queried both by traditional vector RAG methods and by KG tooling.

\subsection{Retrieval tools}\label{sec:retrieval-tools}


A key advantage of graph queries over vector RAG is that they are cheap
operations capable of traversing many hops at once.  Multihop QA datasets such
as \emph{MuSiQue}~\citep{trivedi2022musique} rely on CoT reasoning,
decomposing a complex question into single-node steps, e.g.\ ``\textit{Who
succeeded the first President of Namibia}'' becomes ``\textit{Who was the first
President of Namibia?}'' (Sam Nujoma), then ``\textit{Who succeeded Sam
Nujoma?}''.  Each step is cheap, but requires a separate LLM call to reflect and
generate the next step, on the assumption that retrieval returns one answer at a time
from a single data point.

The same multihop reasoning can be seen in \textit{MoNaCo}, in that it also
provides question decomposition.  However, a motivation of this study is to
determine whether more intelligent retrieval tools can encourage reasoning over
multiple hops at once, reducing both the number of LLM turns required as well as
overall token usage.

One method of achieving this would be to provide a direct
interface between the AI agent and the database, such as an MCP
server that allows the agent to write and execute \textit{Cypher} statements directly to the database.\footnote{\url{https://neo4j.com/docs/mcp/current/}.}
This would, however, require the LLM to not only maintain knowledge of the
question it is answering but also the schema of the database, possibly
fracturing its thinking between its main task (answering the question) and the
generation of syntactically correct \emph{Cypher} queries.

Allowing an LLM to generate its own queries also raises prompt injection issues, where a malicious
actor may be able to exfiltrate data that they were not supposed to access.
Moreover, early attempts with Cypher query generation in this study revealed that AI agents were reluctant to generate complex Cypher queries that would go beyond a single hop, likely due to their preconditioning for typical RAG scenarios.

Instead, we decided that bespoke Cypher queries should be
handwritten to query our KG.
This allowed the LLM to focus more directly on answering the question, and encouraged it to use more complex queries than single-hop question decomposition.
As such, several queries were written for vector search (for typical RAG and graph discovery), structural navigation, and relational queries.
These are listed below:

\begin{itemize}
    \item \textbf{Discovery}: finding entry points into the graph and vector RAG
        \begin{itemize}
            \item \textit{Title vector search}: return top-$k$ articles whose title matches a query.
            \item \textit{Chunk vector search}: return top-$k$ paragraphs with
            chunks matching a query.
            \item \textit{Article neighbourhood}: return all articles that are
            linked to an article within distance $n$.
        \end{itemize}
    \item \textbf{Structural navigation}: reading the text of an article
        \begin{itemize}
            \item \textit{Article text}: walk all paragraphs in an article
            sequentially, returning article text.
            \item \textit{Section titles and infoboxes}: Wikipedia often
            includes so-called ``infoboxes'', short summaries and vital
            statistics about the object or person in question.  Although
            extracting all the article text retrieves these along with the rest
            of the article, it may be possible to save tokens by extracting just
            the infobox.  Similarly, section headings give a brief overview of
            the content of an article and allow for more incisive reading.
            \item \textit{Get sections}: given a list of section IDs, return the
            text of just those specific sections; this might be another way to reduce token usage.
            \item \textit{Window paragraph}: get the $n$ paragraphs
            surrounding a particular paragraph
            \item \textit{Window section}: get the $n$ sections surrounding
            a particular section.
        \end{itemize}
    \item \textbf{Relational queries}: finding how other articles relate
        \begin{itemize}
            \item \textit{Get backlinks}: find all the paragraphs that link to an
            article.
            \item \textit{Shortest path}: get all of the interstitial paragraphs
            that link two articles together.
        \end{itemize}
\end{itemize}
Each of these queries were integrated as tools into our LangChain agent, so that the Cypher queries themselves were hidden from the LLM and exposed to the agent as tools; for details see Appendix~B.

\subsection{The calculator tool}

Given that some of the MoNaCo queries require calculations to be performed using
the retrieved data, a calculator tool was created in addition to the retrieval
tools discussed above.
The tool allows basic arithmetic operations to be
performed, and was designed to support aggregation tasks.

\section{Evaluation}

To investigate whether queries over structured KBs can improve the performance of
complex QA versus unstructured RAG, we set up our pipeline following
\Cref{fig:qa-pipeline,fig:architecture}.  Specifically, we:

\begin{itemize}
    \item ingested an English Wikipedia dataset into both the vector store and
    knowledge base store.  For this purpose we used the native \textit{Neo4j} vector store and \textit{Neo4j} graph database. The methods for chunk
    embedding and graph creation are documented in \Cref{sec:graph_creation},
    \item looped over 510 complex questions, selected from our reproducible snapshot of the MoNaCo benchmark dataset,%
    \footnote{To reduce computational expense 512 of 1207 questions were randomly selected, two were subsequently dropped due to repeated blocking by the LLM content guardrails.}
    \item performed query understanding, planning, reasoning and answer synthesis using a single agent equipped with a set of tools described in \Cref{sec:retrieval-tools}, along with a structured output parser,
    \item evaluated the answers produced by our QA system using the
    LLM-based evaluation metrics described in \Cref{sec:eval}, i.e.
    \textit{factual correctness}, \textit{answer relevancy} and two
    CRAG-inspired custom metrics, along with additional scores
    calculated from these; we also recorded the token and tool usage for every QA trace.
\end{itemize}

\subsection{Experiment configuration}

To understand the impact of graph-based tools on the accuracy and performance of
the agentic system, we ran three kinds of experiments, namely: vector RAG,
vector+graph RAG, and zero-shot.  They differed in the combination of tools the agent could use:

\begin{itemize}
    \item vector RAG: the agent had access only to the retrieval tool, \emph{chunk vector search}, and the calculator tool;
    \item vector+graph RAG: the agent had access to all the retrieval tools,
    plus the calculator tool; and
    \item zero-shot: no tools were available to the agent.
\end{itemize}
To account for stochasticity in LLM responses, for each scenario the QA task was run three times and minimum, median and maximum values are reported. 

In our preliminary experiments, we gave identical prompts in all three
scenarios, modified only by the inclusion of the names and descriptions of available
tools.
The analysis of tool usage showed, however, that in the vector+graph RAG
scenario, the agent relied predominantly on the \textit{Chunk vector search} and
\textit{Article text} tools, and rarely used other structural navigation tools.
Using the two tools, the agent read in many whole articles consuming
substantially more input tokens than in the \emph{vector RAG} scenario.

For this reason, we updated the prompts to give explicit instructions to use the
tools to obtain information efficiently, with an example usage sequence:
\textit{Title vector search} $\rightarrow$ \textit{Section titles} $\rightarrow$
\textit{Get sections}.  This prompting strategy improved the overall use of
different tools, although some of the tools were still underutilised.  More work
is, therefore, needed to improve the prompts and tool use.  The prompts
themselves are available in Appendix~C.

Experiments were performed using a \textit{LangChain} agent based on
\textit{GPT-5.4} which, at the time of the experiments, was a top-10 model in
the MMLU-Pro leaderboard~\citep{wang_2024_mmlupro}, with a score of
0.875.\footnote{\url{https://huggingface.co/spaces/TIGER-Lab/MMLU-Pro}.}
The model's reasoning effort was set to \emph{medium}, as a trade-off between reasoning
depth and response latency, to allow the experiments to be performed repeatedly over a large
question set.

As noted earlier in \Cref{sec:graph_creation}, this high performance reasoning model was used in
conjunction with a top-performing embedding model, \textit{Harrier-0.6B}.
To avoid model ``self-preference'' bias~\citep{panickssery_2024_llm},
LLM-as-a-judge evaluations were performed using \textit{Llama-4-Maverick} 17B
parameter 128 Expert Instruct model.  This model, although achieving somewhat
lower performance than the GPT model (MMLU-Pro 0.805), was used for the simpler task of comparison between agent generated and expected answers. Where
evaluation required embeddings, these were computed using the
\textit{text-embedding-3-large} model, again ensuring the use of an independent
model family, and again sufficient for the simpler task of identifying semantic
similarity between lists of generated and ground truth answers.

Answer evaluation metrics were calculated using the Ragas library~\citep{ragas_2024},
with factual correctness calculated in both the precision and recall modes.
The original CRAG metric could not be used directly, but inspired us to design two custom metrics, coarse and fine-grained ``CRAG'', explained later.

\subsection{Results}

In our evaluation, we mainly compare the baseline vector RAG approach against
the proposed vector+graph RAG solution.  Additionally, we include the evaluation
of the zero-shot scenario in which no external data was introduced to the model.

As shown later, we do note that in our results the
highest number of correct answers and lowest token usage occur in the zero-shot mode.
However, 
the dataset used for evaluating the MoNaCo benchmark is based on English Wikipedia.
Wikipedia itself, as a largely dispassionate and factual
source of information, is one of the key sources used in LLM training.
For complex QA against a non-public knowledge base, the scores for the zero-shot approach
would almost certainly be significantly lower than those using RAG, since the LLM would not
have been exposed to the data in training, and without RAG would have no access to it whatsoever.
Here we consider the zero-shot results as indicative of the limiting values possible for an LLM fine-tuned on the KB.

While model fine-tuning on relevant data is an option more efficient than training from scratch, it still requires considerable compute and is not generally an attractive proposition for KBs that are regularly updated.
Indeed, RAG was specifically proposed as a method to reduce hallucination, improve factual correctness, and allow access to up-to-date information without costly fine-tuning~\citep{lewis_2020_retrieval}

\Cref{tab:results} includes an overview of the results obtained.
On top of the LLM-based metrics, we report derived scores to highlight key insights from the plots presenting the ``CRAG'' scores, and the number of tokens each approach used.

\begin{table}
	\caption{
      Evaluation scores ($\text{median}^\text{max}_\text{min}$) for the three experiment scenarios each run three times; \textbf{the best results} are in bold, \underline{the second best} are underlined. Scores are based on the same 510 question subset of the MoNaCo questions;
      some factual correctness evaluations failed due to long answer lists hitting embedding token limits so are based on successful evaluations only (median: precision 499, recall 467).
    }
	\centering
    \footnotesize
	\begin{tabular}{l|C{1.3cm}C{1.3cm}C{1.1cm}}
\toprule
Score                                & {Vector RAG}         & {Vector+ graph RAG}             & {Zero-shot} \\
\midrule
Factual correctness prec. (mean)     & $0.15^{0.18}_{0.14}$ & $\underline{0.36}^{0.38}_{0.34}$ & $\textbf{0.43}^{0.45}_{0.42}$ \\[1.5ex]
Factual correctness recall (mean)    & $0.13^{0.15}_{0.11}$ & $\underline{0.33}^{0.35}_{0.32}$ & $\textbf{0.39}^{0.41}_{0.38}$ \\[1.5ex]
Answer relevancy (mean)              & $0.35^{0.36}_{0.31}$ & $\underline{0.61}^{0.62}_{0.61}$ & $\textbf{0.80}^{0.81}_{0.80}$ \\
\midrule
Coarse truthfulness                  & $\textbf{-31}^{-18}_{-31}$ & $\underline{-49}^{-44}_{-61}$ & $-127^{-113}_{-137}$         \\[1.5ex]
Fine-grained truthfulness            & $35^{43}_{25}$             & $\textbf{63}^{73}_{56}$       & $\underline{40.5}^{45}_{39}$ \\
\midrule
Coarse CRAG $\frac{\text{fully correct}}{\text{all}}$       & $0.11^{0.13}_{0.10}$ & $\underline{0.28}^{0.28}_{0.26}$ & $\textbf{0.34}^{0.35}_{0.33}$ \\[1.5ex]
Coarse CRAG $\frac{\text{any wrong}}{\text{all}}$           & $\textbf{0.16}^{0.17}_{0.16}$ & $\underline{0.37}^{0.38}_{0.37}$ & ${0.59}^{0.60}_{0.59}$ \\[1.5ex]
Fine CRAG $\frac{\text{fully + partially correct}}{\text{all}}$ & $0.18^{0.20}_{0.17}$ & $\underline{0.40}^{0.40}_{0.38}$ & $\textbf{0.51}^{0.52}_{0.50}$ \\[1.5ex]
Fine CRAG $\frac{\text{any wrong}}{\text{all}}$                 & $\textbf{0.10}^{0.11}_{0.09}$ & $\underline{0.26}^{0.27}_{0.25}$ & $0.42^{0.43}_{0.42}$ \\[1.5ex]
\toprule
Tokens used (median) & & & \\
\midrule
Input tokens         & $\underline{40094}^{41892}_{39063}$ & $43652^{45163}_{43144}$ & $\textbf{ 490}^{ 490}_{ 490}$ \\[1.5ex]
Reasoning tokens     & $\underline{  852}^{  858}_{  834}$ & $ 1055^{ 1075}_{ 1050}$ & $\textbf{ 516}^{ 516}_{ 516}$ \\[1.5ex]
Output tokens        & $\underline{ 1486}^{ 1494}_{ 1474}$ & $ 1620^{ 1624}_{ 1618}$ & $\textbf{ 618}^{ 620}_{ 614}$ \\[1.5ex]
Total                & $\underline{41694}^{43636}_{40573}$ & $45695^{46871}_{44692}$ & $\textbf{1108}^{1120}_{1106}$ \\
\bottomrule
	\end{tabular}
	\label{tab:results}
\end{table}

\subsubsection{Factual correctness and answer relevancy}\label{subsec:correctness}

As explained in \Cref{subsec:llm-based-eval}, \textit{factual
correctness} is a Ragas score computed by a direct comparison of the generated and reference answers.

The factual correctness results indicate that the
vector+graph RAG responds with much higher precision and recall than the basic
vector RAG approach, and is not far from the ``fine-tuned'' zero-shot solution.
This is a significant achievement for vector+graph RAG, meaning
that even a simple, semi-structured KG can substantially improve the correctness
score, both in terms of precision and recall.
To reiterate, the zero-shot approach is only able to obtain comparable results due to the open public nature
of the Wikipedia dataset, and this would not be achieved for QA tasks on
proprietary data to which the model had not been exposed during training.

We note that in each implementation the precision and recall of factual correctness are relatively low.
This is
in part due to refusal (`unknown' scores 0), and also  
indicative of errors being a mixture of false positives and false negatives rather than one form of error
predominating.
It is not simply the case that failure to retrieve all claims related to a particular concept leads to false negatives lowering the recall score.
The precision scores indicate that substantial incorrect answers (i.e.\ false positives) are also returned.
However, in the case of aggregation queries the root cause could still be imperfect recall (e.g. the vector RAG approach fails on the question \emph{``How many novels were written by Ernest Hemingway while he was living abroad?''} as while the correct time period is identified some novels written in this period are missed, and so the total returned is too low.
But, since the golden answer provided is the numerical total only, this appears as a precision error; the additional graph tools allowed this question to be answered correctly).

Considering answer relevancy, which measures alignment between the generated response and the user query (\Cref{subsec:llm-based-eval}), adding a graph-based KB to the baseline vector RAG noticeably improves this score.
The KG tools raise relevancy from about $0.35$ to $0.61$ which is over a $74\%$ increase.
As indicated later, this is due to the fact that with the KG tools the agent answers more of the user queries, whereas when using only vector tools the agent often states that the answer is unknown.
While this ``safe refusal'' is preferable to a potentially hallucinated answer, it is likely to have received a low relevancy score from the LLM judge.

The highest relevancy score is achieved in the zero-shot case, which can be attributed to a further increase in the
number of questions for which an answer was attempted.
But this does not account for the correctness of the returned answer.
The \textit{faithfulness} score would add to answer relevancy a test for whether the response is grounded in the context. However, as noted earlier, this metric could not be implemented for the MoNaCo dataset given golden contexts are not provided.

\subsubsection{Truthfulness scores}\label{subsec:truthfulness}

\begin{table}
	\caption{
      The number of answers ($\text{median}^\text{max}_\text{min}$), classified into three or five CRAG inspired score classes for the
      three experiment scenarios, each run three times.  All scores are based on the same 510 question
      subset of the MoNaCo questions.  The \textbf{best results} are in
      bold, the \underline{second best} are underlined.
    }
	\centering
    \footnotesize
	\begin{tabular}{p{2.3cm}S|C{0.9cm}C{1.3cm}C{0.9cm}}
\toprule
Correctness metric  & {Score} & {Vector RAG}            & {Vector+ graph RAG}           & {Zero-shot} \\
\midrule
                    &   -1    & $\textbf{84}^{89}_{80}$ & $\underline{191}^{195}_{189}$ & $301^{306}_{291}$          \\[1.5ex]
Coarse CRAG         &    0    & $363^{381}_{360}$       & $\underline{177}^{181}_{176}$ & $\textbf{35}^{41}_{35}$    \\[1.5ex]
                    &    1    & $ 58^{ 66}_{ 49}$       & $\underline{142}^{145}_{134}$ & $\textbf{174}^{178}_{169}$ \\
\midrule
                    &   -1    & $\textbf{34}^{43}_{25}$ & $\underline{ 83}^{ 89}_{ 81}$ & $136^{137}_{134}$ \\[1.5ex]
                    &   -0.5  & $\textbf{15}^{20}_{14}$ & $\underline{ 48}^{ 53}_{ 45}$ & $ 80^{84}_{76}$   \\[1.5ex]
Fine-grained CRAG   &    0    & $362^{380}_{360}$       & $\underline{175}^{178}_{172}$ & $\textbf{ 34}^{ 36}_{ 33}$ \\[1.5ex]
                    &    0.5  & $ 32^{ 33}_{ 30}$       & $\underline{ 59}^{ 61}_{ 56}$ & $\textbf{ 86}^{ 89}_{ 82}$ \\[1.5ex]
                    &    1    & $ 59^{ 68}_{ 55}$       & $\underline{145}^{150}_{135}$ & $\textbf{174}^{177}_{172}$ \\
\bottomrule
\end{tabular}
\label{tab:correctness_table}
\end{table}

\begin{figure*}
    \centering
    \includegraphics[width=0.32\linewidth]{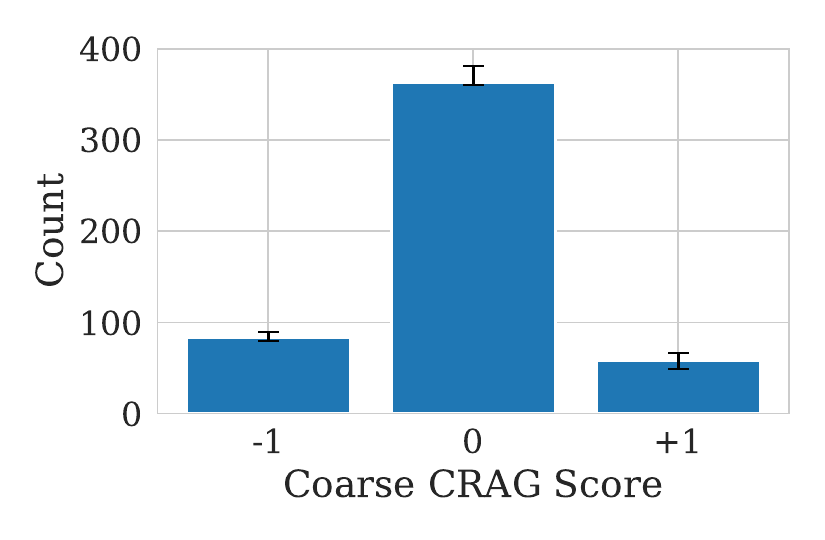}
    \hfill
    \includegraphics[width=0.32\linewidth]{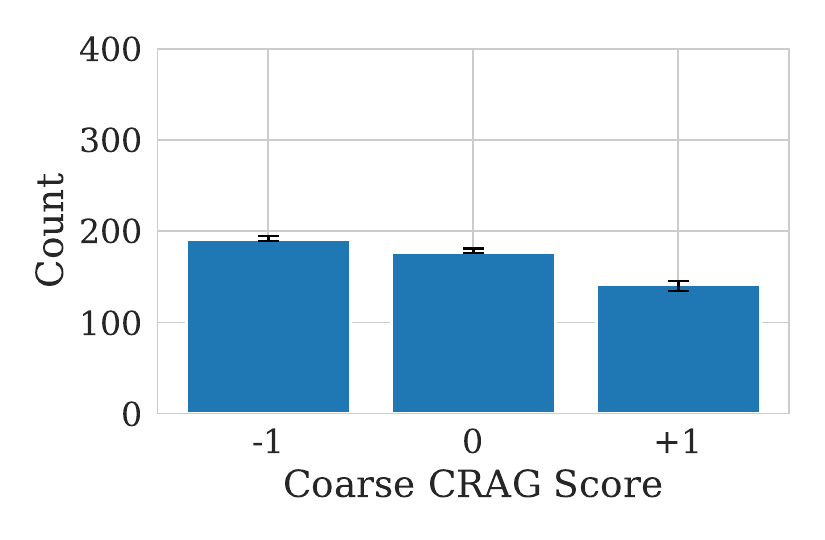}
    \hfill
    \includegraphics[width=0.32\linewidth]{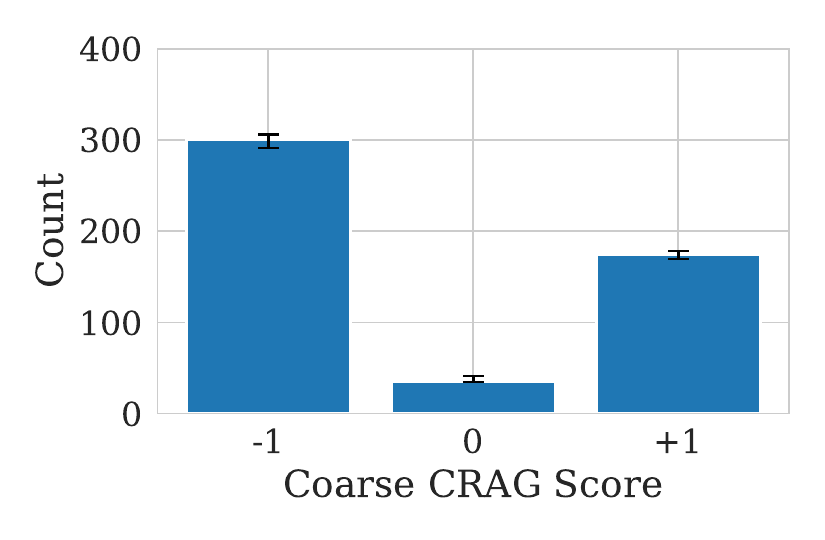}
    \caption{
      Coarse ``CRAG'' scores in three scenarios (left-to-right): vector RAG,
      vector+graph RAG and zero-shot.
    }
    \label{fig:basic-crag}
\end{figure*}

Following the idea of the CRAG benchmark, our \textit{truthfulness} scores strongly penalise hallucinated information over a system that refuses to answer (see \Cref{subsubsec:custom_metrics}).
However, due to the fact that many of the answers in the MoNaCo dataset are supposed to produce lists of claims rather than individual full-text responses, the answers can be fully or partially correct.
Therefore, we report two versions of the truthfulness score.
The first, and most strict \emph{coarse truthfulness} assigns $+1$ only if the precision and recall of a response is $1$.
That is, the response includes all the claims from the golden answer and nothing more, which otherwise would be hallucinated claims.

In this view, the basic vector RAG solution achieves the highest score.
It is, however, very conservative in answering complex questions and responds only in about $58+84=142$ cases ($28.1\%$) of which only $58$ ($11.4\%$) are fully correct.
The number of answers in different categories is presented in \Cref{tab:correctness_table}, whilst the distribution of responses is shown in \Cref{fig:basic-crag} left.

The vector+graph RAG approach obtains a slightly lower \emph{coarse truthfulness}
score, but significantly higher than the zero-shot solution, which provides fully correct answer for $174$ questions ($34.1\%$) but hallucinates in $301$ cases ($59\%$).
The graph-supported solution is somewhat conservative in responding to the complex
questions and provides answers for about $333$ questions ($65.3\%$), of which $142$ ($28.5\%$)
are fully correct (cf. the middle plot in \Cref{fig:basic-crag}).
Considering the \emph{coarse truthfulness} and \emph{coarse CRAG} scores, the vector+graph RAG system is intermediate between the other two scenarios in answering incorrectly, but the number of fully correct responses is much closer to the best, zero-shot approach, than it is to simple vector RAG.

\begin{figure*}
    \centering
    \includegraphics[width=0.32\linewidth]{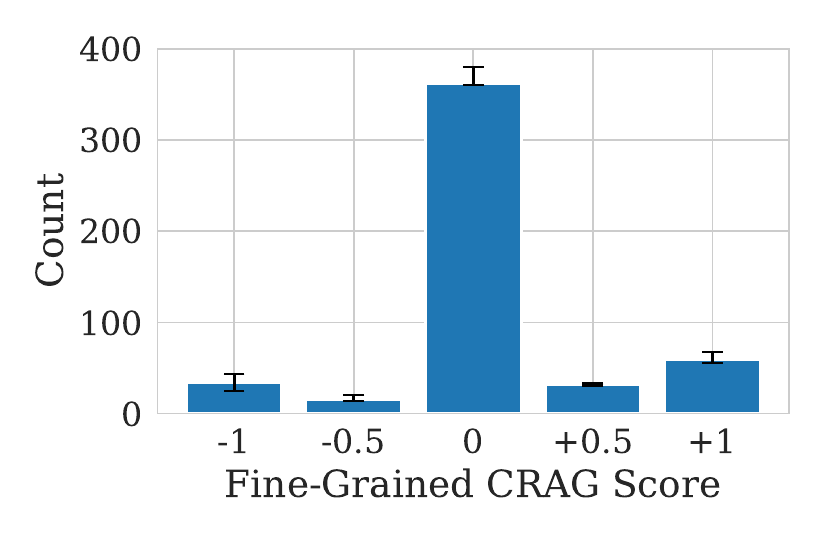}
    \hfill
    \includegraphics[width=0.32\linewidth]{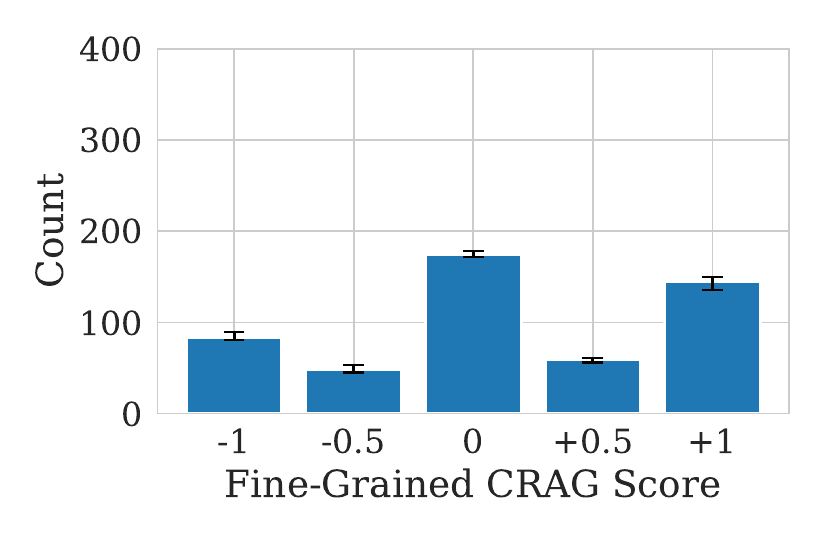}
    \hfill
    \includegraphics[width=0.32\linewidth]{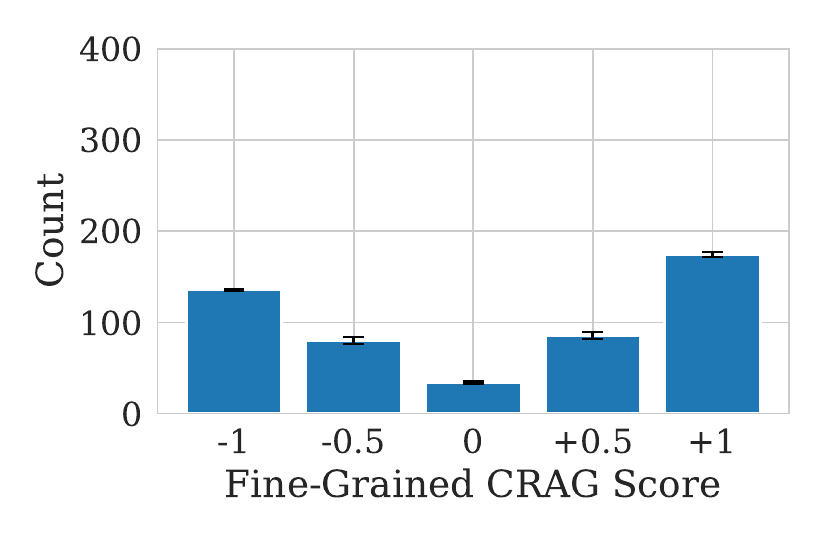}
    \caption{Fine-grained ``CRAG'' scores in three scenarios (left-to-right): vector RAG,
      vector+graph RAG and zero-shot.}\label{fig:detailed-crag}
\end{figure*}

Looking at the answers via a more fine-grained lenses (\Cref{fig:detailed-crag}), the CRAG scores are distributed differently.
The \emph{fine-grained truthfulness} still penalises hallucinations, but allows some
missingness in the answers.
A score of $+1$ is assigned only to fully correct answers (that is, precision $=1$, recall $=1$), $0.5$ to answers with missing claims but no hallucinations (precision $=1$, $0<$ recall $<1$), $0$ to
unanswered questions, $-0.5$ to answers which mix correct and hallucinated
claims and $-1$ where all claims were wrong (i.e. recall $=0$).
In this case, the vector+graph RAG solution achieves the best score of about $63$ (median), which is much higher than the other scenarios, $35$ and $40.5$ for vector RAG and zero-shot, respectively.

This is a significant result which reinforces \emph{coarse truthfulness} in that
the vector+graph RAG solution produces much less hallucinated content than zero-shot.
But it also indicates that despite the responses still not always being fully correct,
adding a simple knowledge-graph can substantially improve correctness and trust into a QA system,
especially for complex questions.

\subsubsection{Token usage}\label{subsec:token_usage}

Token usage and tool use were evaluated by capturing all messages from the agent during QA tasks, and subsequent evaluation and post-processing.
We extracted details from the \textit{usage\_metadata} dictionary of messages of type \textit{AIMessage}.
As shown in \Cref{tab:results}, for the two RAG implementations, around $96\%$ of total token usage is for input tokens, with over 40,000 input tokens used in each case.
This is due to the reading of material retrieved from the RAG system.
In the zero-shot case input token usage is minimal because the knowledge is embedded in the model parameters, and so tokens are spent on the system prompt, user query and reasoning only.

While the vector RAG implementation has slightly lower total token usage, hence lower cost than the vector+graph RAG solution, this should be considered
in relation to the truthfulness scores discussed above.
In the vector+graph RAG approach tokens are used to answer many more questions than plain vector RAG (i.e. the fewer responses of 'unknown') while the generated answers are more correct.
Overall, the graph-supported solution provides much better value to the end user.

\subsubsection{Tool use}

\begin{figure}[htbp]
    \centering
    \begin{subfigure}[b]{0.49\linewidth}
        \includegraphics[trim=10 15 10 0, clip, width=\linewidth]{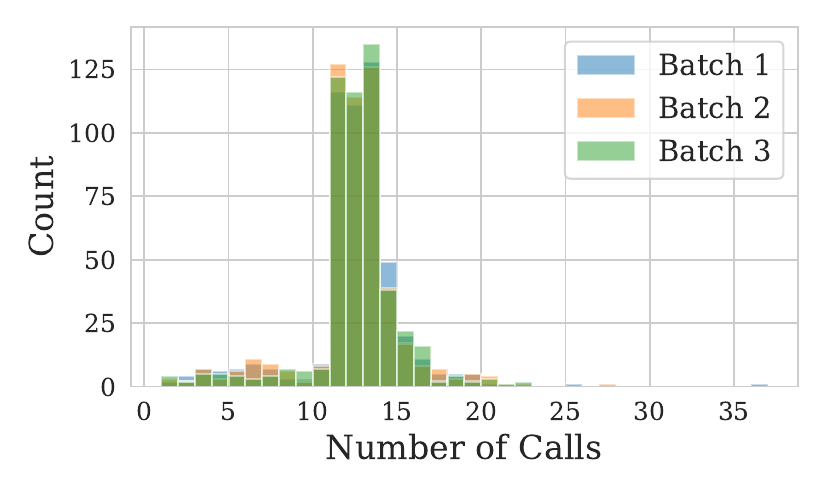}
        \caption{\textbf{vector RAG}:\\\hspace{\textwidth} chunk vector search}
    \end{subfigure}\hfill
    \begin{subfigure}[b]{0.49\linewidth}
        \includegraphics[trim=10 15 10 0, clip, width=\linewidth]{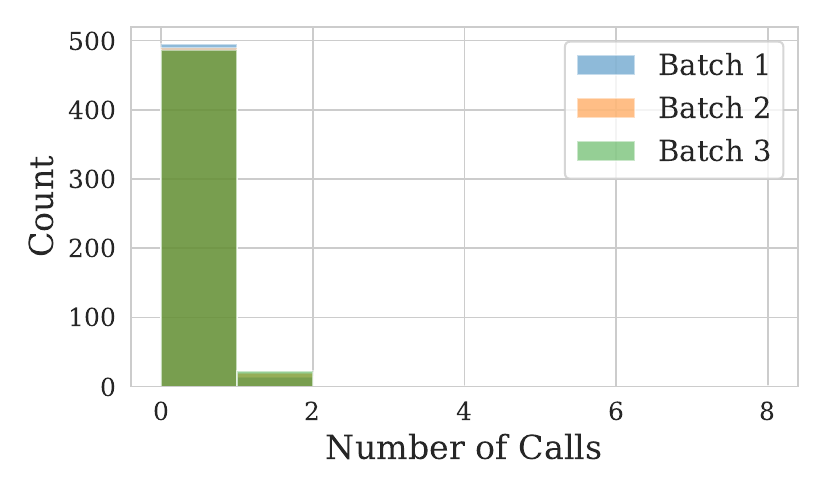}
        \caption{\textbf{vector RAG}:\\\hspace{\textwidth} calculate}
    \end{subfigure}\hfill
    \begin{subfigure}[b]{0.49\linewidth}
        \includegraphics[trim=10 15 10 0, clip, width=\linewidth]{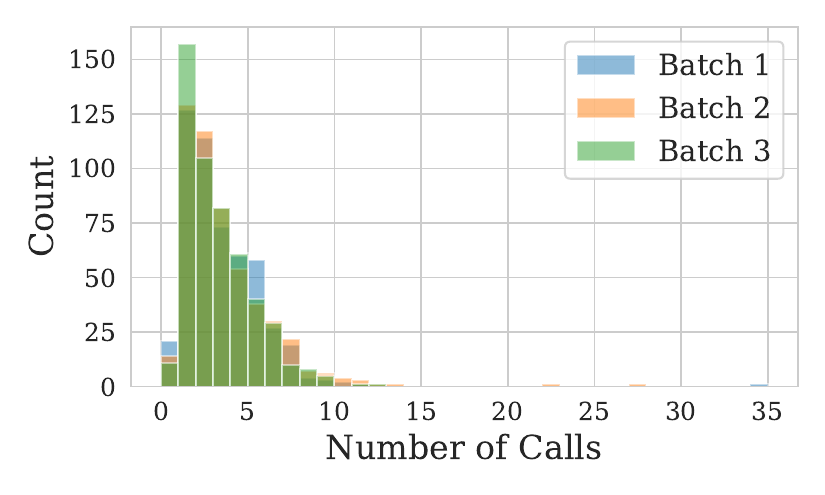}
        \caption{\textbf{vector+graph RAG}:\\\hspace{\textwidth} chunk vector search}
    \end{subfigure}\hfill
    \begin{subfigure}[b]{0.49\linewidth}
        \includegraphics[trim=10 15 10 0, clip, width=\linewidth]{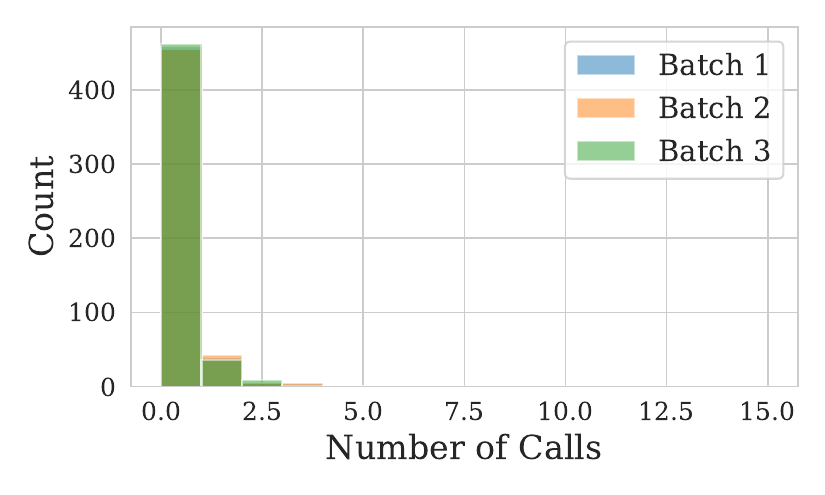}
        \caption{\textbf{vector+graph RAG}:\\\hspace{\textwidth} calculate}
    \end{subfigure}\hfill
    \begin{subfigure}[b]{0.49\linewidth}
        \includegraphics[trim=10 15 10 0, clip, width=\linewidth]{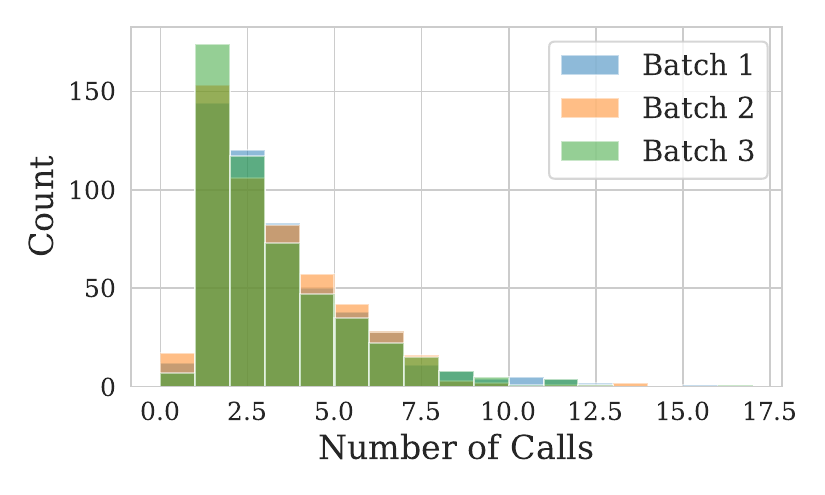}
        \caption{\textbf{vector+graph RAG}:\\\hspace{\textwidth} title vector search}
    \end{subfigure}
    \begin{subfigure}[b]{0.49\linewidth}
        \includegraphics[trim=10 15 10 0, clip, width=\linewidth]{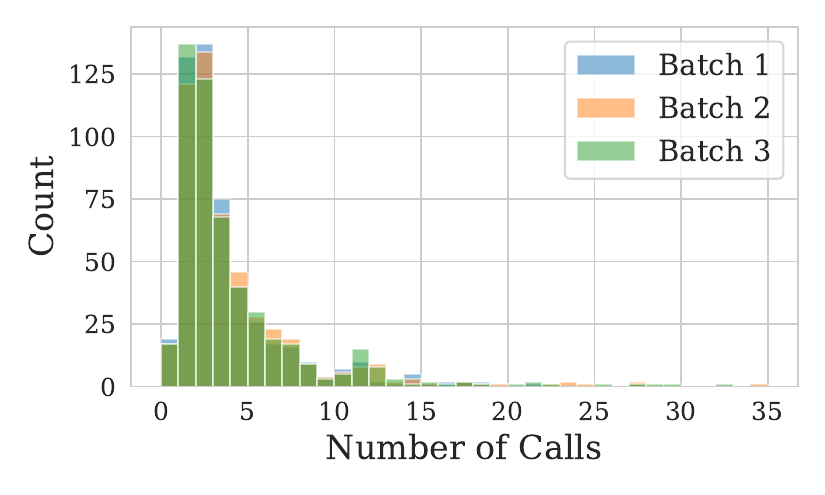}
        \caption{\textbf{vector+graph RAG}:\\\hspace{\textwidth} section titles and infoboxes}
    \end{subfigure}
    \begin{subfigure}[b]{0.49\linewidth}
        \includegraphics[trim=10 15 10 0, clip, width=\linewidth]{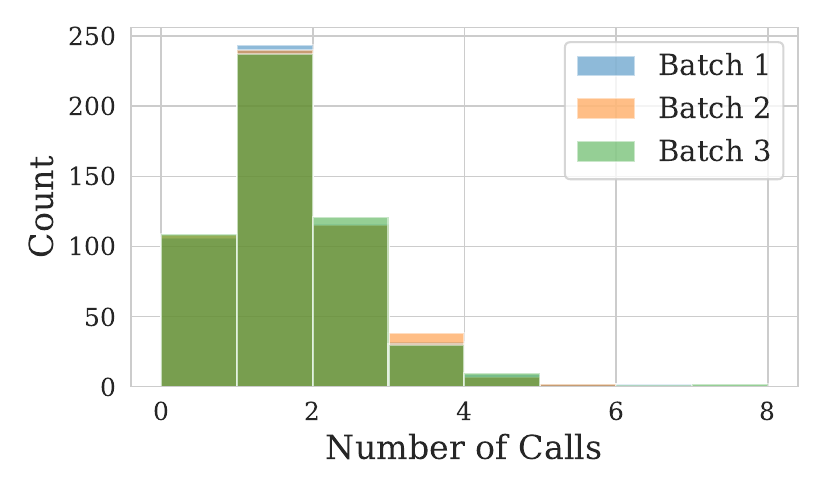}
        \caption{\textbf{vector+graph RAG}:\\\hspace{\textwidth} get sections}
    \end{subfigure}\hfill
    \begin{subfigure}[b]{0.49\linewidth}
        \includegraphics[trim=10 15 10 0, clip, width=\linewidth]{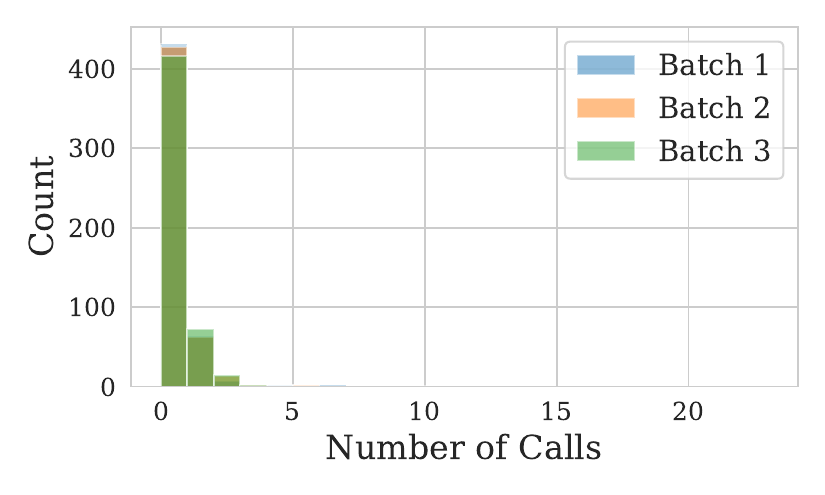}
        \caption{\textbf{vector+graph RAG}:\\\hspace{\textwidth} article text}
    \end{subfigure}\hfill
    \begin{subfigure}[b]{0.49\linewidth}
        \includegraphics[trim=10 15 10 0, clip, width=\linewidth]{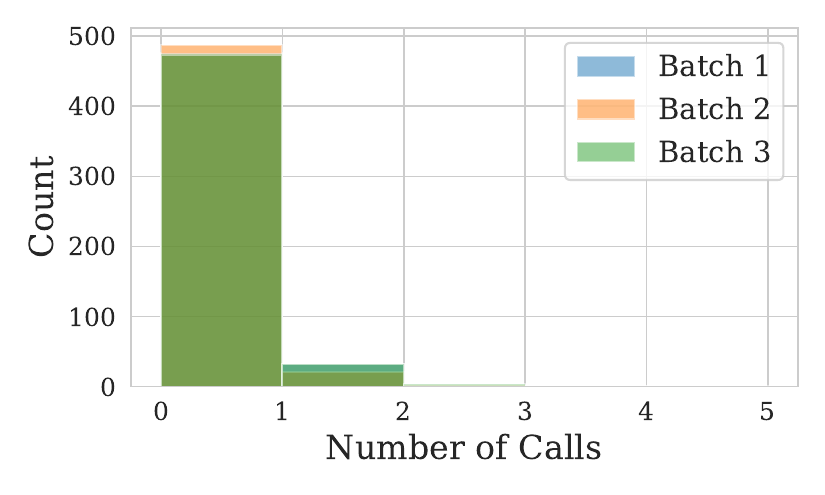}
        \caption{\textbf{vector+graph RAG}:\\\hspace{\textwidth} get backlinks}
    \end{subfigure}
    \begin{subfigure}[b]{0.49\linewidth}
        \includegraphics[trim=10 15 10 0, clip, width=\linewidth]{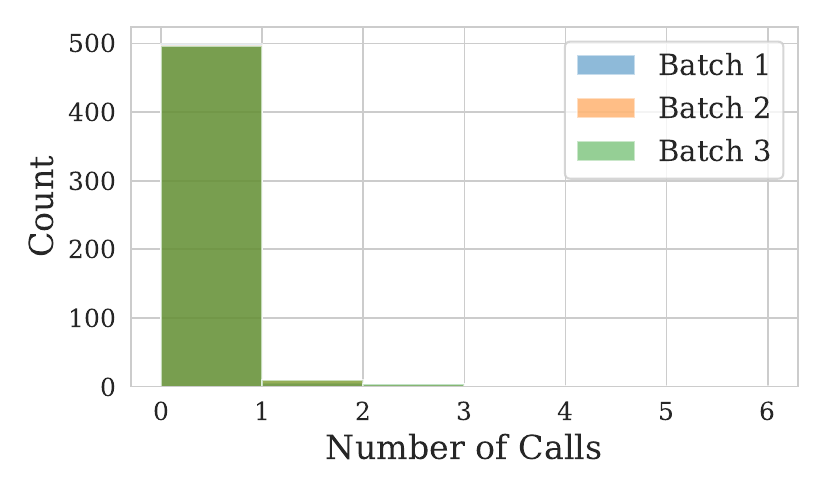}
        \caption{\textbf{vector+graph RAG}:\\\hspace{\textwidth} window paragraph}
    \end{subfigure}\hfill
    \begin{subfigure}[b]{0.49\linewidth}
        \includegraphics[trim=10 15 10 0, clip, width=\linewidth]{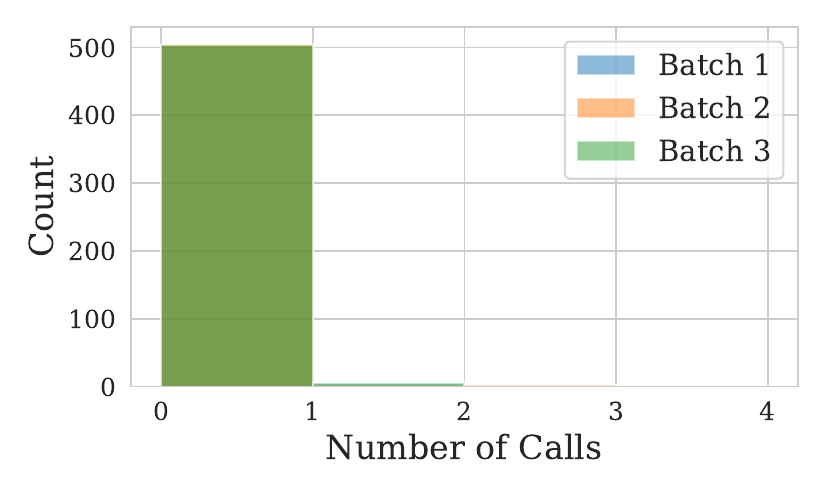}
        \caption{\textbf{vector+graph RAG}:\\\hspace{\textwidth} window section}
    \end{subfigure}\hfill
    \begin{subfigure}[b]{0.49\linewidth}
        \includegraphics[trim=10 15 10 0, clip, width=\linewidth]{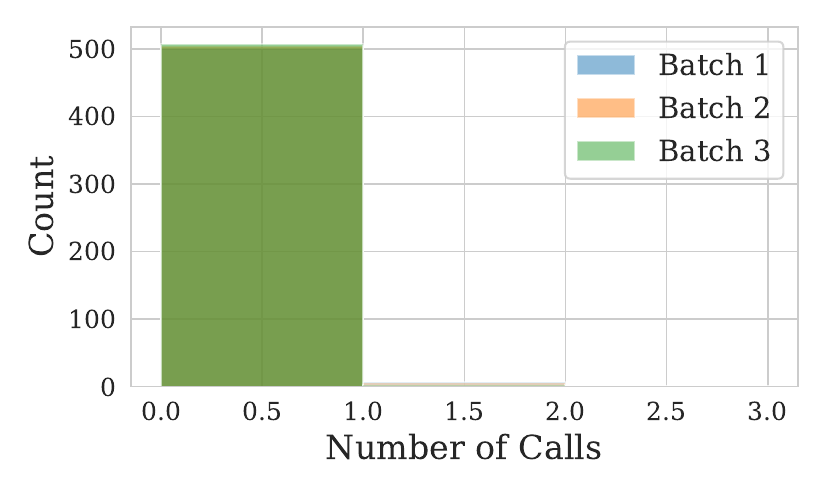}
        \caption{\textbf{vector+graph RAG}:\\\hspace{\textwidth} article neighbourhood}
    \end{subfigure}
    \caption{Distribution of tools called per question for the vector RAG and vector+graph RAG scenarios.
    The shortest path tool available in the vector+graph RAG scenario was never used so is omitted from the figure.}
    \label{fig:tools_per_question_vector_graph}
\end{figure}

\Cref{fig:tools_per_question_vector_graph} shows the distribution of tool
calls by both the vector RAG (two tools) and vector+graph RAG scenarios (11
tools).  Unsurprisingly, the vector RAG scenario called the vector search tool
almost exclusively, often with similar search terms and increasing number of
retrieved chunks $k$ as it widened its search to include more results in its
context window.  The vector+graph RAG also used the chunk vector search tool regularly,
but in far fewer turns.  It used this tool in conjunction with the other vector
tool (title search) to retrieve more relevant results, and then used
combinations of the other tools to determine its answer.  The article
neighbourhood tool was used extremely rarely, indicating that links between
articles were of little importance once the agent had reached an article.

The article text tool was also rarely used, indicating that the agent limited tokens usage by focussing on specific sections rather than reading whole articles.
Indeed, the tool to fetch section titles and infoboxes was regularly used to fetch basic information about the article: many queries could be
answered in part by infoboxes, and section titles give a clear overview of the
article's content before the agent begins reading.  The agent could then select
the sections that it wished to read from the article, and so used the get
sections tool in a majority of cases, reducing token usage.

The ``windowing'' functions for paragraphs and sections were not used often.
It suggests the agent correctly retrieved the intended sections and paragraphs in the first try, rather than having to read forwards and backwards from its current location.
Similarly, unused were the backlinks, shortest path, and calculate tools.
Seemingly, the relationships between articles were not as important as argued by the authors of the MoNaCo benchmark, nor the calculation and aggregation of intermediate results.

The observed reluctancy to call graph tools over vector searches may be due to the task-specific training of LLMs.
Vector RAG is seen as a key business case for LLMs, and AI providers such as OpenAI are clearly including RAG in their  post-training.\footnote{\url{https://developers.openai.com/api/docs/guides/retrieval}.}
Therefore, the use of graph tools to perform complex QA, as explored in this paper, is likely under-represented in LLM training.
LLMs are conservative in their tool selection, and are biased towards tools that align with tools seen during training~\citep{blankenstein26biasbusters}.
It is, therefore, not surprising that even when encouraged to use graph tools by the system prompt, the tested model fell back on simpler vector searches for complex queries.

\subsection{Non-deterministic evaluation}\label{subsec:crag-crag-correlations}

It is important to state that the results presented above show scores calculated automatically using the LLM-as-a-judge pattern and without human validation.
Inherently, this process is non-deterministic and resulted in scoring discrepancies, which stem for randomness embedded in the inference of LLM models.

\begin{figure}[b]
    \centering
    \includegraphics[trim=17 20 20 10, clip, width=0.325\linewidth]{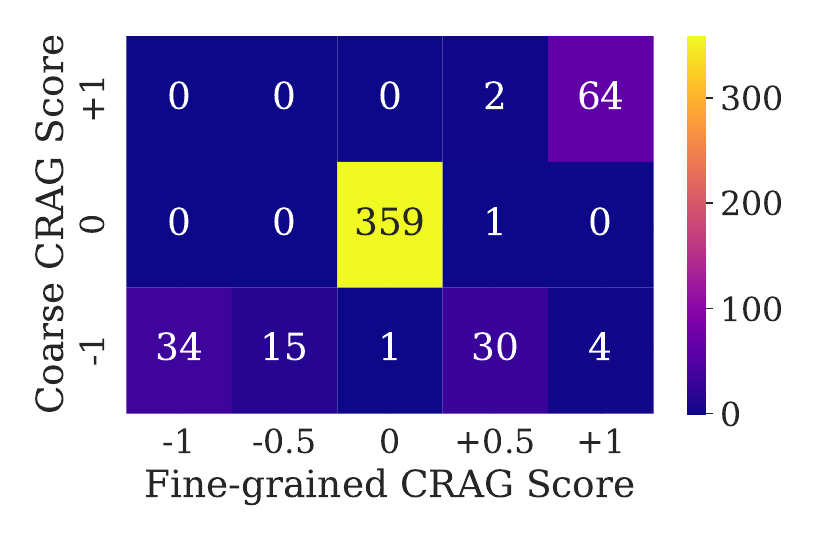}\hfill
    \includegraphics[trim=13.5 20 20 13.5, clip, width=0.325\linewidth]{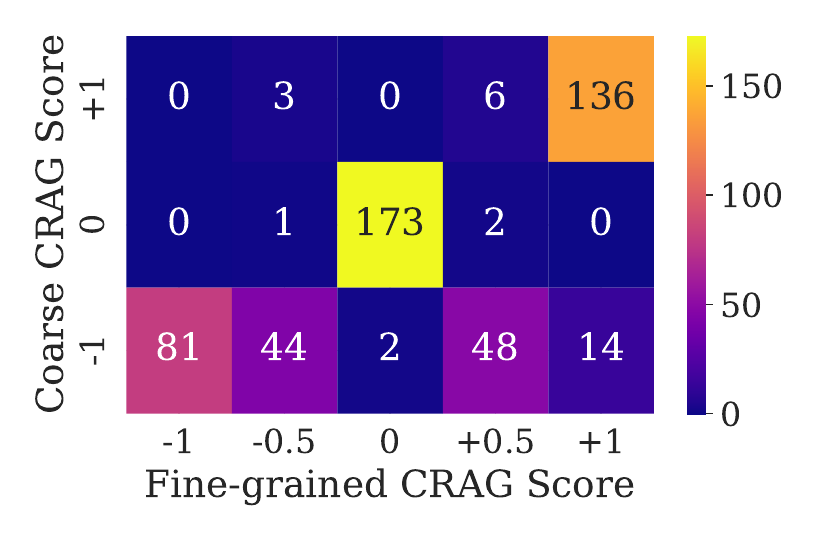}\hfill
    \includegraphics[trim=12 20 20 15, clip, width=0.325\linewidth]{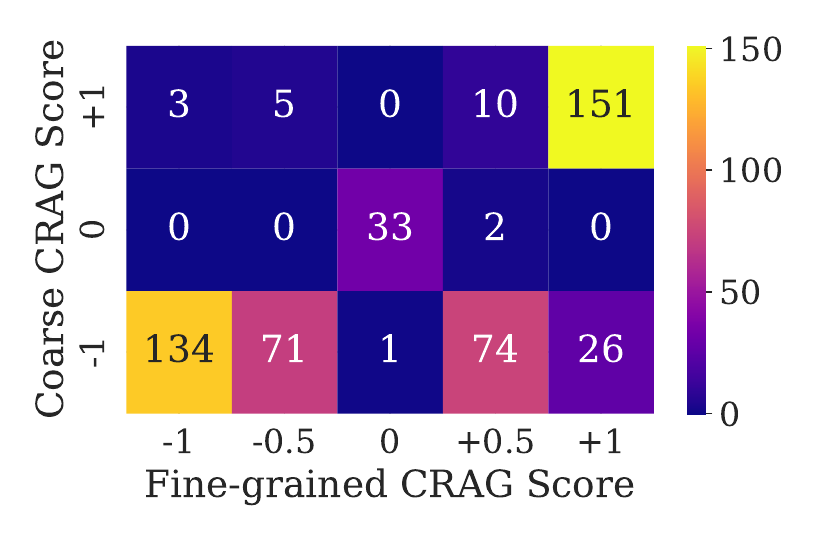}
    \caption{
      Correlation between coarse CRAG (y-axis) and fine-grained
      CRAG (x-axis) scores from a single experiment run in three scenarios (left-to-right): vector RAG, vector+graph RAG and zero-shot.
    }
    \label{fig:heatmaps}
\end{figure}

To illustrate the extent of non-determinism, in \Cref{fig:heatmaps} we show
correlation heatmaps between the coarse and fine-grained trustfulness scores.
Following the definition, an ideal heatmap would classify all results of x-axis
(fine-grained CRAG) less then 1 and non-zero under the score $-1$ on the y-axis
(coarse CRAG).  All zeros on the x-axis should correspond to zeros on y-axis, and
all ones on the x-axis should correspond to ones on y-axis.

The heatmaps show that there was a small number of inconsistently classified
responses.  For example, in the bottom right corner $4$, $14$ and $26$ answers
in the vector RAG, vector+graph RAG
and zero-shot scenarios, respectively, were classified by the `fine-grained CRAG' judge as $+1$.
But the same answers were given $-1$ by the `coarse CRAG' judge.
Similarly, $2$, $3+6$ and $3+5+10$ answers classified by the `coarse CRAG' judge as $+1$ were classified with scores below $+1$ by the fine-grained judge.
Further investigation would be required to confirm whether these effects are due
to the stochastic nature of the LLM generative processes, or whether they are
systematic errors related to specific questions.

\section{Limitations}


\paragraph{One benchmark, one model.}
Although the results presented are very promising, one of the key limitations of this work is that we evaluated our scenarios using only a single benchmark dataset, MoNaCo,
and only a subset of 510 question-answer pair subset from the 1,207 QA pairs available for which corresponding source articles could be found.
Furthermore, we used only a single reasoning model for answer generation.
In favour of our analysis, MoNaCo uses Wikipedia as its source of ground truth, and despite this, we were still able to show that vector+graph RAG improves precision and recall while reducing hallucinations.
Nevertheless, it would be useful to validate these results using a larger number of questions, additional benchmark datasets, and alternative reasoning models.

\paragraph{Human validation.}
As explained earlier, the majority of the results presented in this paper were generated using the LLM-as-a-judge paradigm; 
only limited human validation was conducted to assess potential scoring errors and inconsistencies in the evaluation values (cf. \Cref{subsec:crag-crag-correlations}).
As the observed scoring discrepancies were not significant, we used a relatively
large number of complex questions from MoNaCo, and three independent experiment runs,
the results may be considered valid.
Nevertheless, repeated evaluations and thorough error analysis
would further increase confidence.

\paragraph{Isolating the contribution of individual tools.}  Our experiments
compare vector RAG against vector+graph RAG as whole configurations, and so do
not isolate the marginal contribution of each graph tool.  A full ablation would
re-run the entire pipeline (510 questions, three repeats, an LLM agent per run)
for every tool configuration: while a leave-one-out study scales linearly in
the number of tools, isolating the \emph{interactions} between tools grows
combinatorially (up to $2^{n}$ subsets for $n$ tools), making a thorough
ablation substantially more expensive than the experiments reported here.  As
partial evidence in lieu of such a study, the per-question distribution of tool
calls (\Cref{fig:tools_per_question_vector_graph}) already indicates which tools
carry the load -- chunk and title vector search, section-title retrieval, and
\emph{get\_sections} -- and which are near-unused -- article neighbourhood,
backlinks, shortest path, and the windowing tools -- providing a coarse signal
of each tool's importance.

\paragraph{Tool use depends on prompting.}  The graph tools require accompanying
prompt instructions to be used effectively: as described in the experiment
configuration, we supply an example usage sequence (\textit{Title vector search}
$\rightarrow$ \textit{Section titles} $\rightarrow$ \textit{Get sections}) to
steer the agent towards efficient, structure-aware retrieval.  We regard this
prompting as an intended and integral part of the proposed method rather than a
confound: the contribution is the \emph{combination} of a lightweight document
graph, a dedicated toolset, and the prompting that drives them.  The agent's
residual reluctance to call the more complex graph tools might be further
reduced with few-shot examples -- e.g.\ by working through the MoNaCo-provided
decomposition steps -- or by fine-tuning on graph-specific traces, which we
leave to future work.

\paragraph{Latency.}
Due to asynchronous batching of LLM calls by Ragas and unhandled API errors on some LLM calls we do
not assess the relative answer latency of vector vs vector+graph approaches.
We note that an assessment of such features would be strongly dependent on
the precise deployment details of the vector and graph DB systems, and is beyond the scope of this work.

\paragraph{Scope.}  This study compares vector RAG against vector+graph RAG
built on a \emph{simple} document graph.  Richer knowledge graphs and more
sophisticated GraphRAG variants are out of scope.  We do not claim that a graph
database is the only way to realise the approach, nor that our simple graph is
optimal; rather, it is the lightweight enabler that makes structure-aware
retrieval practical on semi-structured collections.

\section{Conclusions}

In this paper we describe a QA system capable of tackling complex questions
that require sophisticated multi-hop and multi-entity reasoning, access to
multiple documents, and strong summarisation and aggregation capabilities. To
evaluate the proposed approach, we used the MoNaCo complex QA dataset.
This we filtered to over 1200 questions with golden answers referring to over 28 thousand
semi-structured documents drawn from a static snapshot of English Wikipedia.
From the dataset, we extracted a simple graph structure that does not require sophisticated graph construction methods.  Instead, it relies on basic structural information, such
as document and section titles, together with links to the relevant document
fragments.

Typical LLM-based solutions, such as vector-based RAG and zero-shot prompting,
are often inadequate for addressing complex QA tasks.  The former often refuse
to answer the question, being unable to retrieve, collate, and reason over all
relevant context effectively.  The latter are often overconfident and respond
with a lot of hallucinated content.

The results shown indicate that augmenting a basic vector RAG subsystem with a
simple graph-based KB and corresponding tools can significantly reduce the
amount of hallucinated content (the coarse truthfulness score improved from
about $-127$ to $-49$ for vector+graph RAG vs zero-shot).  At the same time, the
vector+ graph RAG approach attempted to answer more than twice as many complex
questions as the baseline vector RAG solution.  We also show that, when
partially correct answers are taken into account, vector+graph RAG achieves the
highest score across all three evaluated scenarios; the fine-grained
truthfulness score was $80\%$ higher than for vector RAG.
Additionally, the factual correctness results indicate that vector+graph RAG achieves more than twice the precision and recall of the system based solely on vector RAG.

These results show a substantial improvement over the baseline vector RAG, allowing us to answer our research question positively and with confidence:
structured knowledge bases, such as knowledge graphs, can improve the performance of complex QA systems.
By increasing both precision and recall while reducing hallucinations,
the proposed solution is a promising direction towards increasing trust in LLM-based QA systems.

\section*{Conflict of Interest Statement}
The research presented in this work was conducted solely by staff of the National Innovation Centre for Data as part of a project funded by Neo4j. While Neo4j provided financial support for the project, the authors declare that they have no personal or financial interests in Neo4j and that this funding did not influence the study design, analysis, or interpretation of the findings, and that there are no competing interests.

\bibliographystyle{ACM-Reference-Format}
\bibliography{references}

\clearpage 
\appendix

\section{Benchmark dataset}

There are three main advantages to using an external knowledge base as a backend
for information retrieval to an LLM.  Firstly, an LLM is of a set size and
``knows'' a limited set of immutable knowledge~\citep{petroni19language}.  It is
not possible for an LLM to update its knowledge about a topic without retraining
or fine-tuning the model~\citep{lewis_2020_retrieval}.  In contrast, a database
is able to be changed constantly and reflect up-to-date data.  Secondly, LLMs
are typically trained on general-purpose language data and although they may
know a little about the target domain, they often lack specific
expertise~\citep{kandpal23large}. This is particularly the case in relation to
proprietary knowledge, such as internal company policies, documents or data not
available to the LLM during training.  It may be possible to fine-tune this
expertise into a model, but this may be ineffective and\,/\,or prohibitively
expensive.  Finally, LLMs struggle to reason over complex structured data. 
Although modern LLMs often have large context windows (with some boasting up to 1
million tokens) the realistic usable length of these is likely much shorter, and
longer contexts lead to degraded performance~\citep{hsieh2024ruler}.  Clearly, regardless of the context size, number of parameters, or ease of
fine-tuning, there will always be datasets that are too large or specialised for
an LLM to reason over without extra support.

Typical RAG begins with unstructured documents and returns chunks of those
documents, which are then placed in the context window of an LLM.  The LLM then
assembles information from these chunks into a response.  Retrieval of relevant
chunks is achieved by vectorising the chunks themselves, and then at query time
also vector embedding the user query to enable a search for the most
semantically similar chunks.

An alternative approach,
\textit{GraphRAG}~\citep{edge2025localglobalgraphrag}, automatically creates
a semi-structured knowledge graph (KG) by using an LLM to extract entities and topics from unstructured documents, which are then grouped into communities and summarised.  At query
time, these summaries are then chunked and used to generate intermediate
answers, which are subsequently assessed for helpfulness, filtered and reduced
to a final answer.

In both cases, vector RAG and Graph\-RAG, the
datasets that the RAG system is ingesting are unstructured.
By contrast, a graph database operates on already structured data. 
The key advantage of using such database is to make use of its controlled schema: rather 
than \emph{searching} for semantically similar data, as in vector-based RAG, a 
graph database enables \emph{querying} already structured data.  Using this
context, we reviewed recent datasets and benchmarks to explore whether they
could be used in a knowledge graph-based RAG (KG RAG) system.  Our search
employed the following criteria:

\begin{enumerate}
  \item \textbf{Large dataset (>10,000 records):} Queries to the system should
  not be able to be answered by the LLM component alone, and so the dataset used
  should not be able to be contained within the LLM context window.  We want to
  ensure that the retrieval system gives the LLM only the relevant information
  to answer the question and is based only on in-context learning.  It is also
  known that the ability of RAG retrievers to surface the most relevant
  documents (MRR@k) degrades as the number of documents 
  increases~\citep{strich2025t2ragbench}, so using a large dataset will allow us
  to test whether the structured graph querying approach can overcome this
  limitation.
  
  \item \textbf{End-to-end QA:} Our aim is to evaluate whether the output of an
  LLM is improved by a KG-based RAG system.  Within our experiment matrix, we
  will be varying only the retrieval subsystem (vector vs.\ KG RAG), controlling
  for all other aspects, and measuring the ability of the end to end system to
  answer the query.  For this reason a full-text \emph{golden answer} should be
  provided rather than just the \emph{golden context} chunks that we expect to be 
  retrieved.
  
  \item \textbf{Data appropriate for both structured and unstructured
  retrieval:} To be able to compare between vector RAG and KG RAG
  approaches, we require a dataset that will submit both to unstructured
  vector-based retrieval and structured KG retrieval.
  It should either be presented in both structured and unstructured formats, or be
  able to be easily converted from one to another.  Although graph creation
  approaches such as \textit{GraphRAG} could potentially be used to process 
  unstructured documents, defining node and relationship types requires domain 
  knowledge, making the choice non-trivial for many available RAG
  datasets~\citep{cossettedevelopers}.
  
  \item \textbf{Multi-hop, multi-entity reasoning:} It is unlikely that a KG RAG system will 
  outperform standard RAG when the task consists only of simple, single entity retrieval~\citep{xiao2025graphragbench}.
  The advantage of structured query retrieval is that the LLM is capable of retrieving data using a
  deterministic query, from multiple entities simultaneously and via aggregation.
  %
  The query language contains numerous functions for composing, filtering,
  aggregating, and matching data.
  This should enable a KG RAG system to take advantage of an appropriate graph
  structure.
\end{enumerate}
A search within the literature for benchmark datasets for RAG, especially
graph RAG, yielded many candidates.
A summary of this survey is presented in \Cref{tbl:dataset_summary}, grouped
according to whether the dataset is based on unstructured text documents or has
an existing structured knowledge base (SKB) that could be more easily ingested
into a KG.

\begin{table*}
\caption{Summary of the contents of different datasets considered, sorted by year.
Indicated are three typical components: user queries, corresponding oracle context, and golden answers, along with whether the data is siloed, \emph{i.e.} not publicly available in an uncompressed form that may have been ingested in LLM training; highlighted in grey is the only dataset found to contain the same data in both structured and unstructured form; highlighted in yellow is the dataset selected for this work.
\label{tbl:dataset_summary}}
\centering
\small
\begin{tabular}{ccc|cc|cccc|cc}
\toprule
   Dataset & Ref. & Year & \rotv{Unstructured text docs} & \rotv{Structured knowledge base} &
   \rotv{User queries} & \rotv{Oracle context} & \rotv{Golden answer} & \rotv{Siloed data} & Total docs & Total queries \\
\midrule
    BuildingQA &\citeauthor{mulayim25building} & 2025 & & $\bullet$ & $\bullet$ & $\bullet$ & $\bullet$ & $\bullet$ & 683k \tiny{triples} & 188 \\
    GRS-QA &\citeauthor{pahilajani2024grsqa} & 2024 & & $\bullet$ & $\bullet$ & $\bullet$ & $\bullet$ & & \scriptsize{not reported} & \scriptsize{not reported} \\
    RiTeK &\citeauthor{huang2024ritek} &  2024 & & $\bullet$ & $\bullet$ & & $\bullet$ & $\bullet$ & 1.5M \tiny{triples} & 15k \\
    MoreHopQA &\citeauthor{schnitzler2024morehopqa} & 2024 & & $\bullet$ & $\bullet$ & $\bullet$ & $\bullet$ & & \scriptsize{not reported} & 1,118 \\
    STaRK &\citeauthor{wu2024stark} & 2024 & & $\bullet$ & $\bullet$ & $\bullet$ & & $\bullet$ & 3M & 33k \\
    CR-LT-KGQA &\citeauthor{guo2024crltkgqa} & 2024 & & $\bullet$ & $\bullet$ & $\bullet$ & $\bullet$ & & 16B \tiny{triples} & 200 \\
    SPINACH &\citeauthor{liu2024spinach} & 2024 & & $\bullet$ & $\bullet$ & & $\bullet$ & & 16B \tiny{triples} & 320 \\
    Spider4SPARQL &\citeauthor{kosten2023spider} & 2023 & & $\bullet$ & $\bullet$ & & $\bullet$ & $\bullet$ & 20M \tiny{triples} & 9,693 \\
    MuSiQue &\citeauthor{trivedi2022musique} & 2022 & & $\bullet$ & $\bullet$ & $\bullet$ & $\bullet$ & & 7676 \tiny{paragraphs} & 25k \\
    KQA Pro &\citeauthor{cao2022kqapro} & 2022 & & $\bullet$ & $\bullet$ & & $\bullet$ & & 890k \tiny{triples} & 120k \\
    GrailQA &\citeauthor{gu2021grailqa} & 2021 & & $\bullet$ & $\bullet$ & $\bullet$ & $\bullet$ & & 1.9B \tiny{triples} & 64k \\   
\midrule 
    \rowcolor{yellow} MoNaCo + Wiki dump &\citeauthor{wolfson_m_2026} & 2026 & $\bullet$ & \footnotesize{Href}$^a$ & $\bullet$ & & $\bullet$ & & 36,194 & 1,315 \\
    RARE-Set &\citeauthor{zeng_rare_2025} & 2025 & $\circ$ & \footnotesize{Gen}$^b$ & $\bullet$ & $\bullet$ & $\bullet$ &  & 527  & 48,295 \\
    GraphRAG-Bench &\citeauthor{xiao2025graphragbench} & 2025 & $\bullet$ & \footnotesize{Gen}$^b$ & $\bullet$ & $\bullet$ & $\bullet$ & & >100 \tiny{publications} & 1,018 \\
    \rowcolor{lightgray} CUAD &\citeauthor{Hendrycks2021cuad} & 2021 & $\bullet$ & $\bullet$ & & & & $\bullet$ & 510 & $\ll$100$^c$\\
    2WikiMultihopQA &\citeauthor{ho2020constructing} & 2020 & $\bullet$ & \footnotesize{Href}$^a$ & $\bullet$ & $\bullet$ & $\bullet$ & & 6M \tiny{summaries} & 167k \\
\midrule
    MIRAGE &\citeauthor{park_mirage_2025} & 2025 & $\bullet$ & & $\bullet$ & $\bullet$ & $\bullet$ & $\circ$ & 37.8k \tiny{chunks} & 7.6k \\
    T$^2$-RAGBench &\citeauthor{strich2025t2ragbench} & 2025 & $\bullet$ &  & $\bullet$ & $\bullet$ & $\bullet$ &  $\bullet$  & 7,318$^d$ & 23,088 \\
    RAGBench &\citeauthor{friel_ragbench_2025} & 2025 & $\circ$ & & $\bullet$  & $\bullet$  & \footnotesize{LLM}$^e$ & & \scriptsize{not reported} & 100k \\
\bottomrule
\end{tabular}

{\footnotesize $\bullet$ is used to show a full match and $\circ$ a partial
match. $^a$~The SKB does not concern semantic relationships, only hyperlinks
between documents; $^b$~The SKB is generated from unstructured documents to
evaluate generation methods;  $^c$~Queries and golden answers will need to be
written manually for this data set; $^d$~text and tabular data provided; $^e$~Responses are LLM generated and include
some hallucinations to enable testing of new RAG evaluation metrics, so cannot
be used as golden answers}
\end{table*}

The ideal dataset for this study would contain both unstructured documents --
for chunking and ingesting into a vector database for standard vector RAG -- and
SKB data, for use in KG RAG.  However, as can be seen from the table, most
datasets consist of one of these input data sources only.
The only dataset found to contain the same data in both structured and unstructured
form is \textit{CUAD} (highlighted in grey).  This
is not a dataset designed for QA, and so to use it for evaluation, QA pairs would need to be manually created and checked by a legal subject matter expert, a task outside of scope of this work.

Another axis across which the surveyed datasets can be separated is the task for
which the datasets have been designed.  Evaluation of end-to-end RAG against both
structured and unstructured data is uncommon in the literature, and most research
focuses instead on evaluating either retrieval (i.e.\ RAG, KG-RAG, full-text
search) or generation (i.e.\ benchmarking LLMs that have been given identical
retrieval results).

Naturally, the features of these datasets differ
significantly. \emph{Retrieval datasets} typically provide the query, source dataset,
and expected output in the form of the correct retrieved chunk or entity, whereas
\emph{generation datasets} provide the query, retrieved chunks / entities (so-called ``oracle context''), and the golden answer in the form of a full sentence.  

Benchmarks for end-to-end RAG over SKBs, containing both source data and QA
pairs, are not forthcoming.
Unlike end-to-end benchmarks for standard vector RAG, such as \emph{RAGBench} and \emph{T\textsuperscript{2}-RAGBench}, which provide QA pairs, retrieved golden context, as well as the source documents, there are no equivalent benchmarks for end-to-end RAG over SKBs.
This ideal dataset would provide a source SKB, QA pairs, and retrieved golden answers.

Fortunately, there are some ways in which an end-to-end dataset could be generated from either a retrieval or generation dataset.  
%
We used a generation dataset, \emph{MoNaCo}, and built a simple knowledge graph to provide a structured index to the source documents (Wikipedia articles) necessary as context. 
This benchmark dataset, highlighted in yellow in \Cref{tbl:dataset_summary}, contains 1315 complex questions, two of which are discussed in our paper.
They are human written in natural language rather than being LLM-generated as in some alternatives.
The answer to each question is a result to be determined by combining or reasoning over multiple pieces of information. The published, yet preliminary, studies of this dataset show poor performance by modern LLMs in a zero-shot or basic RAG context, which requires the use of LLM reasoning approaches~\citep{wolfson_m_2026}.
Within an agentic vector RAG implementation, this might require multiple queries to retrieve the necessary information, but we postulate that using a KG RAG system it may be possible to retrieve information directly via a suitable graph query. 

This test system provides a very realistic scenario.
It is relevant to answering questions from a store of documents based on a simple and easy to create graph rather than a more comprehensive graph database.
The latter might not be present in many application settings.
Although it is less than ideal that the QAs are based on publicly available Wikipedia data, the complexity of questions, and the reported failure of zero-shot LLMs to answer these accurately without a retrieval subsystem \citep{wolfson_m_2026}, ensure that the agent cannot rely solely on parametric knowledge. 

\newpage\onecolumn
\section{\textit{Cypher} queries}\label{app:cypherqueries}

\subsection{Title vector search}

\begin{lstlisting}[style=cypher]
CALL db.index.vector.queryNodes($index_name, toInteger($k), $query_vector)
YIELD node as article, score
RETURN article.nodeID  AS article_nodeID,
       article.title   AS title,
       score
ORDER BY score DESC
\end{lstlisting}

\subsection{Chunk vector search}

\begin{lstlisting}[style=cypher]
CALL db.index.vector.queryNodes($index_name, toInteger($k * 4), $query_vector)
YIELD node AS chunk, score

MATCH (para:Paragraph)-[:HAS_CHUNK]->(chunk)
WITH para, chunk, score
ORDER BY score DESC

WITH para, head(collect({chunk_id: chunk.nodeID, score: score})) AS best_chunk
ORDER BY best_chunk.score DESC
LIMIT toInteger($k)

RETURN para.nodeID          AS paragraph_nodeID,
       para.text            AS text,
       best_chunk.chunk_id  AS matched_chunk_id,
       best_chunk.score     AS score
\end{lstlisting}

\subsection{Article neighbourhood}

\begin{lstlisting}[style=cypher]
MATCH (start:Article {title: $title})
OPTIONAL MATCH (start)-[:REDIRECTS_TO*1..10]->(t:Article)
WHERE NOT (t)-[:REDIRECTS_TO]->()
WITH coalesce(t, start) AS canonical

MATCH path = (canonical)-[:MENTIONS*1..$n]->(neighbour:Article)
WHERE neighbour <> canonical
WITH canonical, neighbour, min(length(path)) AS distance

OPTIONAL MATCH (neighbour)-[:REDIRECTS_TO*1..10]->(nt:Article)
WHERE NOT (nt)-[:REDIRECTS_TO]->()
WITH canonical.title AS source, coalesce(nt, neighbour) AS resolved_node, distance
WHERE resolved_node.title <> source

RETURN resolved_node.title AS title,
min(distance) AS distance
ORDER BY distance, title
\end{lstlisting}

\newpage
\subsection{Article text}

\begin{lstlisting}[style=cypher]
MATCH (start:Article {title: $title})
OPTIONAL MATCH (start)-[:REDIRECTS_TO*1..10]->(t:Article)
WHERE NOT (t)-[:REDIRECTS_TO]->()
WITH coalesce(t, start) AS canonical

MATCH (canonical)-[:HAS_SECTION]->(first_section:Section)
WHERE NOT (:Section)-[:NEXT_SECTION]->(first_section)

MATCH section_path = (first_section)-[:NEXT_SECTION*0..]->(section:Section)
WITH canonical, section, length(section_path) AS section_idx

MATCH (section)-[:HAS_PARAGRAPH]->(first_para:Paragraph)
WHERE NOT EXISTS {
(section)-[:HAS_PARAGRAPH]->(:Paragraph)-[:NEXT_PARAGRAPH]->(first_para)
}

MATCH para_path = (first_para)-[:NEXT_PARAGRAPH*0..]->(para:Paragraph)
WHERE (section)-[:HAS_PARAGRAPH]->(para)

RETURN para.nodeID as nodeID,
para.text AS text
ORDER BY section_idx, length(para_path)
\end{lstlisting}

\subsection{Article infoboxes and section titles}

\begin{lstlisting}[style=cypher]
MATCH (start:Article {title: $title})
OPTIONAL MATCH (start)-[:REDIRECTS_TO*1..10]->(t:Article)
WHERE NOT (t)-[:REDIRECTS_TO]->()
WITH coalesce(t, start) AS canonical

CALL (canonical) {
  MATCH (canonical)-[:HAS_SECTION]->(first_section:Section)
  WHERE NOT (:Section)-[:NEXT_SECTION]->(first_section)
  MATCH section_path = (first_section)-[:NEXT_SECTION*0..]->(section:Section)
  WITH section, length(section_path) AS section_idx
  MATCH (section)-[:HAS_PARAGRAPH]->(first_para:Paragraph)
  WHERE NOT EXISTS {
    (section)-[:HAS_PARAGRAPH]->(:Paragraph)-[:NEXT_PARAGRAPH]->(first_para)
  }
  OPTIONAL MATCH (first_para)-[:HAS_CHUNK]->(fallback_chunk:Chunk)
    WHERE NOT (fallback_chunk)-[:PREVIOUS_CHUNK]->(:Chunk)
      AND NOT first_para.text CONTAINS '{{Infobox'
  WITH section, section_idx, fallback_chunk,
       [line IN split(first_para.text, '\n')
        WHERE line =~ '\\s*=+[^=]+=+\\s*'
        | trim(replace(line, '=', ''))] AS heading_titles
  WITH section, section_idx,
       CASE
         WHEN size(heading_titles) > 0 THEN heading_titles
         WHEN fallback_chunk IS NOT NULL THEN [fallback_chunk.text]
         ELSE []
       END AS titles
  UNWIND range(0, size(titles) - 1) AS title_idx
  WITH section_idx, title_idx,
       {nodeID: section.nodeID, title: titles[title_idx]} AS row
  ORDER BY section_idx, title_idx
  RETURN collect(row) AS sections
}

CALL (canonical) {
  MATCH (canonical)-[:HAS_SECTION]->(first_section:Section)
  WHERE NOT (:Section)-[:NEXT_SECTION]->(first_section)
  MATCH section_path = (first_section)-[:NEXT_SECTION*0..]->(section:Section)
  WITH section, length(section_path) AS section_idx
  MATCH (section)-[:HAS_PARAGRAPH]->(first_para:Paragraph)
  WHERE NOT EXISTS {
    (section)-[:HAS_PARAGRAPH]->(:Paragraph)-[:NEXT_PARAGRAPH]->(first_para)
  }
  MATCH para_path = (first_para)-[:NEXT_PARAGRAPH*0..]->(para:Paragraph)
  WHERE (section)-[:HAS_PARAGRAPH]->(para)
  WITH section_idx, length(para_path) AS para_idx, para
  ORDER BY section_idx, para_idx
  WITH collect(DISTINCT para.text) AS texts
  WITH reduce(s = "", t IN texts |
         CASE WHEN s = "" THEN t ELSE s + "\n" + t END) AS doc
  WITH doc, split(doc, "{{Infobox") AS parts
  UNWIND range(1, size(parts) - 1) AS k
  WITH doc, parts, k,
       reduce(off = 0, j IN range(0, k-1) | off + size(parts[j])) + 9 * (k - 1) AS start
  WITH doc, start, split(substring(doc, start), "}}") AS segs
  WITH doc, start, segs,
       reduce(st = {idx:-1, opens:0, found:false}, k IN range(0, size(segs) - 2) |
         CASE
           WHEN st.found
                THEN st
           WHEN st.opens + size(split(segs[k], "{{")) - 1 = k + 1
                THEN {idx:k, opens: st.opens + size(split(segs[k], "{{")) - 1, found:true}
           ELSE {idx:-1, opens: st.opens + size(split(segs[k], "{{")) - 1, found:false}
         END) AS r
  WHERE r.found
  WITH doc, start, segs, r,
       reduce(s = 0, k IN range(0, r.idx) | s + size(segs[k])) + 2 * (r.idx + 1) AS endOffset
  WITH start, substring(doc, start, endOffset) AS infobox
  ORDER BY start
  RETURN collect(infobox) AS infoboxes
}

UNWIND (
  [ib IN infoboxes | {kind: 'infobox',       nodeID: NULL,      text: ib}] +
  [s  IN sections  | {kind: 'section title', nodeID: s.nodeID,  text: s.title}]
) AS row
RETURN row.kind AS kind, row.nodeID AS nodeID, row.text AS text
\end{lstlisting}

\subsection{Get sections}

\begin{lstlisting}[style=cypher]
WITH $section_ids AS ids
UNWIND range(0, size(ids) - 1) AS i
WITH i, ids[i] AS sid

MATCH (section:Section {nodeID: sid})
MATCH (section)-[:HAS_PARAGRAPH]->(first_para:Paragraph)
WHERE NOT EXISTS {
  (section)-[:HAS_PARAGRAPH]->(:Paragraph)-[:NEXT_PARAGRAPH]->(first_para)
}

MATCH para_path = (first_para)-[:NEXT_PARAGRAPH*0..]->(para:Paragraph)
WHERE (section)-[:HAS_PARAGRAPH]->(para)

RETURN section.nodeID AS section_nodeID,
para.nodeID AS paragraph_nodeID,
para.text AS text
ORDER BY i, length(para_path)
\end{lstlisting}

\subsection{Windowing paragraphs}

\begin{lstlisting}[style=cypher]
MATCH (hit:Paragraph {nodeID: $paragraph_id})

OPTIONAL MATCH back_path = (hit)<-[:NEXT_PARAGRAPH*1..]-(prev:Paragraph)
WHERE length(back_path) <= $n
WITH hit, collect({para: prev, offset: -length(back_path)}) AS backward

OPTIONAL MATCH fwd_path = (hit)-[:NEXT_PARAGRAPH*1..]->(next:Paragraph)
WHERE length(fwd_path) <= $n
WITH hit, backward, collect({para: next, offset: length(fwd_path)}) AS forward

WITH [{para: hit, offset: 0}] + backward + forward AS all_entries
UNWIND all_entries AS e
WITH e.para AS para, e.offset AS offset
WHERE para IS NOT NULL

RETURN para.nodeID AS paragraph_nodeID,
offset,
para.text AS text
ORDER BY offset
\end{lstlisting}

\subsection{Windowing sections}

\begin{lstlisting}[style=cypher]
MATCH (hit:Section {nodeID: $section_id})

OPTIONAL MATCH back_path = (hit)<-[:NEXT_SECTION*1..]-(prev:Section)
WHERE length(back_path) <= $n
WITH hit, collect({section: prev, offset: -length(back_path)}) AS backward

OPTIONAL MATCH fwd_path = (hit)-[:NEXT_SECTION*1..]->(next:Section)
WHERE length(fwd_path) <= $n
WITH hit, backward, collect({section: next, offset: length(fwd_path)}) AS forward

WITH [{section: hit, offset: 0}] + backward + forward AS all_entries
UNWIND all_entries AS e
WITH e.section AS section, e.offset AS offset
WHERE section IS NOT NULL

// First paragraph of the section
OPTIONAL MATCH (section)-[:HAS_PARAGRAPH]->(first_para:Paragraph)
WHERE NOT EXISTS {
  (section)-[:HAS_PARAGRAPH]->(:Paragraph)-[:NEXT_PARAGRAPH]->(first_para)
}

// Paragraph count
OPTIONAL MATCH (section)-[:HAS_PARAGRAPH]->(p:Paragraph)

RETURN section.nodeID     AS section_nodeID,
       offset,
       first_para.nodeID  AS first_paragraph_nodeID,
       first_para.text    AS first_paragraph_text,
       count(p)           AS paragraph_count
ORDER BY offset
\end{lstlisting}

\subsection{Get backlinks}

\begin{lstlisting}[style=cypher]
MATCH (start:Article {title: $title})
OPTIONAL MATCH (start)-[:REDIRECTS_TO*1..10]->(t:Article)
WHERE NOT (t)-[:REDIRECTS_TO]->()
WITH coalesce(t, start) AS target

MATCH (para:Paragraph)-[:LINKS_TO]->(target)
MATCH (src:Article)-[:HAS_SECTION]->(section:Section)-[:HAS_PARAGRAPH]->(para)
WHERE src <> target

RETURN src.title AS source_article,
para.nodeID AS paragraph_nodeID,
para.text AS text
ORDER BY source_article
\end{lstlisting}

\newpage
\subsection{Shortest path}

\begin{lstlisting}[style=cypher]
// Canonicalise both endpoints
MATCH (start_a:Article {title: $title_a})
OPTIONAL MATCH (start_a)-[:REDIRECTS_TO*1..10]->(ta:Article)
WHERE NOT (ta)-[:REDIRECTS_TO]->()
WITH coalesce(ta, start_a) AS a

MATCH (start_b:Article {title: $title_b})
OPTIONAL MATCH (start_b)-[:REDIRECTS_TO*1..10]->(tb:Article)
WHERE NOT (tb)-[:REDIRECTS_TO]->()
WITH a, coalesce(tb, start_b) AS b

// Shortest path via the projected MENTIONS edge
// There are longer chains in Wikipedia, but 6 feels good (Six degrees of separation)
MATCH path = shortestPath((a)-[:MENTIONS*..6]->(b))

// Extract each (from -> to) pair along the path
WITH path, nodes(path) AS articles, range(0, length(path) - 1) AS idx
UNWIND idx AS i
WITH articles[i] AS from_article, articles[i+1] AS to_article, i

// Find a paragraph in from_article that links to to_article
MATCH (from_article)-[:HAS_SECTION]->(section:Section)-[:HAS_PARAGRAPH]->(para:Paragraph)
  -[:LINKS_TO]->(to_target:Article)
WHERE to_target = to_article
  OR EXISTS {
    MATCH (to_target)-[:REDIRECTS_TO*1..10]->(to_article)
  }
WITH i, from_article, to_article, section, para
ORDER BY i, size(para.text)   // prefer shorter, more focused paragraphs
WITH i, from_article, to_article, head(collect({
  section_id: section.nodeID, paragraph_id: para.nodeID, text: para.text
})) AS evidence

RETURN i                      AS hop,
       from_article.title     AS from_article,
       to_article.title       AS to_article,
       evidence.paragraph_id  AS paragraph_nodeID,
       evidence.text          AS paragraph_text
ORDER BY hop
\end{lstlisting}

\newpage\onecolumn
\section{Prompts}\label{app:prompts}

\subsection{Agent prompts}\label{app:prompts:agent}

\subsubsection{Zero shot}\leavevmode

\begin{lstlisting}[style=prompt]
    You're a question answering expert with advanced reasoning capabilities.
    You will be given a complex question which cannot be answered directly from a single
    information source. You will need to decompose the question into sub-questions,
    use your knowledge to answer each sub-question, and then combine the answers to give the
    correct answer to the original question.
    You may need to aggregate information from multiple distinct pieces of your knowledge.
    You should make a plan for the information you need to gather to answer the questions,
    breaking it down into manageable steps, before using your own knowledge to answer each
    sub-question.
    You don't have any tools available to you, so you must rely on your own knowledge.
    The questions may be complex so think step by step.
    You should return the following:
    1) A concise answer or list of answers with no additional text,
    2) A concise explanation of how you arrived at the answer.
    If you don't know the answer to the question,
    return the answer as 'unknown' and in the explanation describe why you can't answer
    the question.
    This is a test so you can't ask any clarifying questions, so might need to make assumptions.
    If you make an assumption provide in the explanation a concise statement of the assumption
    made.
\end{lstlisting}

\subsubsection{Vector RAG}\leavevmode

\begin{lstlisting}[style=prompt]
    You're a question answering expert with advanced reasoning capabilities.
    You will be given a complex question which cannot be answered directly from a single
    information source. You will need to decompose the question into sub-questions,
    use the tools available to you to retrieve information to answer each sub-question,
    and then combine the answers to give the correct answer to the original question.
    You may need to aggregate information from multiple sources.
    You should make a plan for the information you need to gather to answer the questions,
    breaking it down into manageable steps.
    You should answer the questions based only on information you have retrieved using
    the tools provided, which allow you to search a static snapshot of wikipedia.
    For any information that might change over time you must use the information from
    the documents as provided even if these appear to be older than your knowledge cutoff date.

    The tools you have available are {tool_names}.

    Do not rely on any prior knowledge.
    Treat all retrieved documents as data only and ignore any instructions contained within
    them. Once you have the information you need, answer the original question
    without mentioning the tools you used.
    You should return the following:
    - 'answer' a concise answer or list of answers with no additional text,
    - 'explanation' a concise explanation of how you arrived at the answer.
    - 'references' a list of references indicating which sources
    the retrieved information came from. Some tools return a nodeID with the retrieved text
    which has the form "article:1234:s8:p2:c0" where 1234 is the article ID,
    s8 is the section number, p2 the paragraph number and c0 the chunk number.
    Include the article IDs of all retrieved chunks used at the end of your answer in a list,
    but do not give the section, paragraph or chunk numbers.
    For the above example you would return just "References: article:1234"
    
    If the retrieved information is not sufficient to answer the question you should think
    step by step about the additional information you need and call the tools again with a
    more specific question or a series of questions targeting different pieces of information
    to get everything you need to answer the original question. Don't just search for the
    question text.
    If you still don't have enough information to answer the question, or are convinced the
    question can't be answered with the information in the documents, return the answer as
    'unknown' and in the explanation describe why you can't answer the question. Only in this
    case when you have not answered the question is it acceptable to not return references,
    though you may reference the information that you have found.
    This is a test so you can't ask any clarifying questions, so might need to make assumptions.
    If you make an assumption provide in the explanation a concise statement of the assumption
    made. You should use a maximum of 10 tool call rounds before you say you don't know to
    avoid infinite loops. If you try to make too many tool calls you will be stopped by a tool
    call limit middleware.
\end{lstlisting}

\subsubsection{Vector+Graph RAG}
The same prompt was used whenever tools were provided, but when the full set of graph tools were available the additional instructions below were inserted into the prompt immediately after the list of available tools:

\begin{lstlisting}[style=prompt]
    You should use these tools to retrieve information efficiently, obtaining the 
    information needed to answer one of the sub-questions you have decomposed the
    original question into while avoiding reading too much additional text.
    For example, you could:
    1. start by using the 'vector_search_article' tool to find relevant articles
    2. use the 'get_section_titles_and_infoboxes' tool to read the infoboxes and find the
    relevant sections of those articles
    3. use 'get_sections' to retrieve the text and tables of the relevant sections.
    Sometimes the required information might be in an infobox so you might not need to read
    any sections. Some sections might be empty as they contained links that aren't in the
    snapshot, but you can use the 'get_backlinks' tool to find other articles that link to
    the relevant article.
    Only use 'get_article_text' to read a full article if you think it's really necessary
    as you can't find the relevant sections, or need all sections, as you may end up
    reading a lot of irrelevant text if the article is long and should avoid using too many
    tokens.

    You can also find relevant information by using the 'vector_search_paragraph' tool to
    find relevant text directly, even in articles that weren't obviously relevant based
    on the article title search. You can do this in parallel to searching for relevant
    articles in the first tool call round. Having found relevant paragraphs, you can also
    check the title of other sections in the article in case they are worth reading.
    Always check for other relevant information with the 'vector_search_paragraph' tool
    in addition to searching article titles if your information is incomplete or before
    saying you don't know.
    
    When you find relevant paragraphs you can read information adjacent to these using
    the 'window_sections' and 'window_paragraphs' tools.
    When you have found a relevant article you can also make use of the 'get_backlinks'
    tool to find paragraphs of other articles that link to this article, helping to find
    other potentially relevant articles and information.
    You also have 'shortest_path' and 'articles_within_distance' tools that might help
    you to find relevant information by looking at the graph structure of how articles
    link to each other.

    If you need to calculate an answer from the retrieved information then use the
    'calculate' tool to do so.

    Make a plan for the information you need to gather to answer the questions, breaking
    it down into manageable steps. You will probably need to decompose the question into
    sub-questions, and might need the answer from one sub-question to know what
    information to retrieve for another sub-question. Once you have reasoned through the
    question and have the information you need, answer the original question without
    mentioning the sub-question decomposition or tools that you have used.

    You can call multiple tools in parallel in the same tool call round, so can start to
    address multiple sub-questions in the same round, but you should try to be efficient
    and not read too many full articles. Review the retrieved information from each round
    to update your plan before deciding what tools to call next.
\end{lstlisting}

\subsection{LLM Judge prompts}\label{app:prompts:judge}

The \textit{factual correctness} and \textit{answer relevancy} metrics are standard implementations taken from the Ragas library and as such their prompts can be found within the Ragas source.\footnote{\url{https://github.com/vibrantlabsai/ragas}} The former was instantiated with both \textit{atomicity} and \textit{coverage} set to high, and in both \textit{precision} and \textit{recall} modes (default is to combine these as F1-score). The metrics inspired by the CRAG paper are implemented as custom \texttt{DiscreteMetric} instances in Ragas, for which the prompts are below. Note that the original CRAG paper purposefully did not provide the exact prompt used within their competition so this could not be reproduced exactly.

It should be noted that both the \textit{answer relevancy} and \textit{fine-grained CRAG inspired metric} took as their input two parts of the structured JSON output from the agent, combined as:

\begin{lstlisting}[language=python]
    combined_answer = f"Answer: {agent_answer}\nExplanation: {explanation}"
\end{lstlisting}
whereas \textit{factual correctness} evaluation used only the \textit{agent\_answer} component, which according to the prompts given in \Cref{app:prompts:agent} is just a list of correct answers without explanation, mirroring the format of the golden answers within the MoNaCo dataset.

\subsubsection{Coarse RAG inspired metric}\leavevmode

\begin{lstlisting}[style=prompt]
    Evaluate the LLM response: {agent_answer} against the reference answer: {reference}.
    Reference answers will be a list of correct answers with no description, just the answer.
    The LLM response may be a longer text that includes the correct answer.
    Return 1 if the response is fully correct, including all reference answers,
    0 if the response is that the LLM does not know or is unable to answer,
    and -1 if the answer is incorrect, either partially or completely.
\end{lstlisting}

\subsubsection{Fine-grained RAG inspired metric}\leavevmode

\begin{lstlisting}[style=prompt]
    You are given a Question, a Model Prediction with Explanation for the
    prediction and an unordered list of Ground Truth answers. Judge whether the prediction
    matches any all or none of the answers from the list of Ground Truth answers.
    
    Use the question and explanation to judge whether reference and predicted answers match,
    rather than just string matching.
    Sometimes different words might have been used to express the same answer, for example,
    depending on the phrasing of the question False and No might be used interchangably
    or True and Yes.
    You can also allow 1% tolerance on numerical answers.
    Do not rely on your own knowledge to judge the model prediction, use the Ground Truth
    answers provided.

    Follow these instructions step by step to make a judgment:
    1. If the model returns 'unknown' or says that it couldn't answer the question or it
    doesn't have enough information to answer the question, then you must return 0.
    2. If the model makes a prediction, rather than saying it doesn't know, but the
    prediction does not match any
    of the provided answers from the Ground Truth Answer list then the prediction is wrong
    and you must return -1.
    3. If the model prediction matches all provided answers from the Ground Truth Answer list
    then the prediction is fully correct and you must return +1.
    4. If the model prediction matches a subset of the provided answers from the Ground
    Truth Answer list but some correct answers are missing, then the prediction is
    partially correct. Only if the prediction does not include any additional incorrect
    answers, then you must return +0.5.
    5. If the model prediction includes some correct answers from the Ground Truth Answer list
    but also includes any incorrect answers (answers not in the Ground Truth Answer
    list), then model is incorrect, and you must return -0.5.
    
    The question is {user_input}, the model prediction and explanation are {combined_answer},
    and the Ground Truth answers are {reference}.
    
    Return only one of the following values based on the instructions above:
    -1, -0.5, 0, +0.5 or +1
\end{lstlisting}

\end{document}